%% file: neurips_2026.tex
\pdfoutput=1
\documentclass{article}

\PassOptionsToPackage{numbers, compress}{natbib}

\usepackage[main, final]{neurips_2026}

\usepackage[utf8]{inputenc}
\usepackage[T1]{fontenc}
\usepackage{hyperref}
\usepackage{url}
\usepackage{booktabs}
\usepackage{amsfonts}
\usepackage{nicefrac}
\usepackage{microtype}
\usepackage{xcolor}

\usepackage{graphicx}
\usepackage{subcaption}
\usepackage{multirow}
\usepackage{colortbl}
\usepackage{xspace}
\usepackage{enumitem}
\usepackage{amsmath}
\usepackage{amssymb}
\usepackage{mathtools}
\usepackage{amsthm}
\usepackage{algorithm}
\usepackage{algpseudocode}
\usepackage{dsfont}
\definecolor{realbrickred}{rgb}{0.8, 0.25, 0.33}
\definecolor{brickred}{rgb}{0.0, 0.45, 0.81}
\definecolor{midnightblue}{rgb}{0.1, 0.1, 0.44}
\definecolor{oceanboatblue}{rgb}{0.0, 0.47, 0.75}
\definecolor{goldenrod}{HTML}{E6B800}

\input{commands}
\title{\name: Free-Form Tabular Data Generation via Joint Numerical–Language Diffusion}

\author{
    \textbf{Donghong Cai$^{1}$ \quad
    Jiarui Feng$^{1}$ \quad
    Yanbo Wang$^{2}$ \quad
    Da Zheng$^{3}$} \\
    \textbf{Yixin Chen$^{1}$ \quad
    Muhan Zhang$^{2}$} \\
    \texttt{\{cai.d, feng.jiarui, ychen25\}@wustl.edu, wangyanbo@stu.pku.edu.cn}\\
    \texttt{zhengda.zheng@antgroup.com, muhan@pku.edu.cn}\\
    $^{1}$Washington University in St. Louis \quad
    $^{2}$Peking University \quad
    $^{3}$Ant Group
}

\begin{document}
\maketitle

\input{00-abstract}
\input{01-introduction}
\input{02-methods}

\input{03-relatedwork}
\input{04-experiments}
\input{06-conclusion}

\begin{ack}
Donghong Cai, Jiarui Feng, and Yixin Chen are supported by NSF grant CBET 2225809, and a Washington University Transcend Initiative grant. Yanbo Wang and Muhan Zhang are supported by the National Natural Science Foundation of China (62550138, 62276003). The authors declare no competing interests.
\end{ack}

\bibliographystyle{ACM-Reference-Format}
\bibliography{07-references}

\newpage
\input{08-appendix}

\newpage
\input{09-checklist}

\end{document}

%% file: commands.tex
\newcommand{\name}{\textsc{TabDLM}\xspace}
\newcommand{\redbf}[1]{\textcolor{brickred}{\textbf{#1}}}

\usepackage[normalem]{ulem}

\newcommand{\names}{\textsc{TabDLM}$_{\text{170M}}$\xspace}
\usepackage{tabularx}
\usepackage{tcolorbox}
\usepackage{float}
\usepackage{titletoc}
\tcbuselibrary{skins,breakable}

\usepackage[normalem]{ulem}
\usepackage{pifont}
\definecolor{exbadc}{HTML}{C62828}
\definecolor{exminc}{HTML}{E08A00}
\tcbset{exrow/.style={enhanced, colback=white, boxrule=0.6pt, arc=1.2mm, left=2mm, right=2mm, top=1mm, bottom=1.2mm,
          toptitle=0.5mm, bottomtitle=0.5mm, fonttitle=\footnotesize\bfseries, coltitle=white, fontupper=\footnotesize,
          before skip=5pt, after skip=5pt},
        exours/.style={exrow, colframe=brickred, colbacktitle=brickred},
        exourslight/.style={exrow, colframe=brickred!45, colbacktitle=brickred!18, coltitle=brickred!75!black},
        exbase/.style={exrow, colframe=black!62, colbacktitle=black!62}}
\newcommand{\exbad}[1]{{\setlength{\fboxsep}{1pt}\colorbox{exbadc!14}{\textcolor{exbadc}{\boldmath\textbf{#1}}}}}
\newcommand{\exminor}[1]{{\setlength{\fboxsep}{1pt}\colorbox{exminc!18}{\textcolor{exminc}{\boldmath\textbf{#1}}}}}
\newcommand{\exbadtxt}[1]{{\color{exbadc}\uwave{#1}}}
\newcommand{\exok}{\textcolor{brickred}{\ding{51}}\hspace{0.4em}}
\newcommand{\exno}{\textcolor{exbadc}{\ding{55}}\hspace{0.4em}}
\newcommand{\exmin}{\textcolor{exminc}{\ding{115}}\hspace{0.4em}}
\newcommand{\exfields}[6]{{\scriptsize\noindent\setlength{\tabcolsep}{1pt}%
  \begin{tabular*}{\linewidth}{@{\extracolsep{\fill}}cccccc@{}}
  \textcolor{black!60}{\texttt{submission\_time}} & \textcolor{black!60}{\texttt{submission\_year}} &
  \textcolor{black!60}{\texttt{title\_length}} & \textcolor{black!60}{\texttt{abstract\_length}} &
  \textcolor{black!60}{\texttt{category}} & \textcolor{black!60}{\texttt{arxiv\_code}}\\
  #1 & #2 & #3 & #4 & #5 & #6
  \end{tabular*}\par}\vspace{3pt}}
\newcommand{\extitle}[1]{\noindent\textcolor{black!60}{\texttt{title}:} #1\par\vspace{1pt}}
\newcommand{\exabstract}[1]{\noindent\textcolor{black!60}{\texttt{abstract}:} #1\par}
\newcommand{\exnotes}[1]{\vspace{3pt}{\scriptsize\noindent #1\par}}
\newcommand{\exsep}{\tcbline}

%% file: 00-abstract.tex
\begin{abstract}
Synthetic tabular data generation has attracted growing attention due to its importance for data augmentation, foundation models, and privacy. However, real-world tabular datasets increasingly contain \emph{free-form text} fields (e.g., reviews or clinical notes) alongside structured numerical and categorical attributes. Generating such heterogeneous tables with joint modeling of \textbf{different modalities} remains challenging. Existing approaches broadly fall into two categories: diffusion-based methods and LLM-based methods. Diffusion models can capture complex dependencies over numerical and categorical features in continuous or discrete spaces, but extending them to open-ended text is nontrivial,
since they model each column as a continuous value or one of a closed set of categories, whereas text lives in an exponentially large sequence space.
In contrast, LLM-based generators naturally produce fluent text, yet their discrete tokenization can distort precise or wide-range numerical values, hindering accurate modeling of both numbers and language. In this work, we propose \name, a unified framework for free-form tabular data generation via a joint numerical--language diffusion model built on masked diffusion language models (MDLMs). \name models textual and categorical features through masked diffusion, while modeling numerical features with a continuous diffusion process through learned specialized numeric tokens embedding; bidirectional attention then captures cross-modality interactions within a single model. Extensive experiments on diverse benchmarks demonstrate the effectiveness of \name compared to strong diffusion- and LLM-based baselines. Our code is available at \url{https://github.com/ilikevegetable/TabDLM}.
\end{abstract}

%% file: 01-introduction.tex
\section{Introduction}
\label{sec:intro}
\raggedbottom
Tabular data is one of the most fundamental data types in modern machine learning and is indispensable across domains such as finance~\cite{sattarov2023findiff}, healthcare~\cite{mimic}, and social media~\cite{lakkaraju2013s}. In practice, however, deploying machine learning on tabular datasets faces recurring obstacles, including privacy and security constraints~\cite{hernandez2022synthetic, assefa2020generating}, limited data availability~\cite{fonseca2023tabular}, and missing values~\cite{zheng2022diffusion,you2020handling}. These challenges motivate synthetic tabular data generation, which seeks to produce synthetic samples that match the schema and key statistical properties of the original dataset for replacing the original data.

High-fidelity synthetic tabular data generation remains challenging because real-world tables exhibit complex dependencies across columns~\cite{xu2019modeling, GreaT} and often contain a mixture of numerical, categorical, and free-form textual features~\cite{robinson2024relbench}. Existing approaches can broadly be categorized into diffusion-based and large language model (LLM)-based methods. Diffusion-based methods employ diffusion processes~\cite{NCSN, DDPM, DDIM, D3PM} to learn denoising transformations that approximate the data distribution during training and to generate realistic samples from noise during inference. Such models have been widely applied to tabular data generation, particularly for continuous (numerical) features~\cite{kim2022sos, kimstasy, zheng2022diffusion}. More recent work has explored combining continuous and discrete diffusion to jointly model numerical and categorical attributes~\cite{zhang2024mixedtype, tabddpm, lee2023codi, tabdiff}. However, extending standard diffusion models to open-ended textual fields remains challenging due to the \textbf{exponential size of the text space}.
Existing tabular diffusion models~\cite{tabddpm, zhang2024mixedtype, tabdiff} decode every non-numerical cell into one of a fixed set of categories, so they cannot generate open-vocabulary text. In contrast, LLM-based methods naturally support free-form text generation~\cite{GreaT, difflm} owing to their strong language modeling capabilities. However, their token-level representations can be unreliable for modeling high-precision or wide-range numerical values~\citep{yang2024number}, as \textbf{numbers are often fragmented into multiple tokens}. Furthermore, \textbf{autoregressive generation enforces a left-to-right dependency structure} that is misaligned with the mutually dependent relationships across tabular columns.

To address the above limitations, we propose \name, a unified framework that integrates the complementary strengths of diffusion-based and LLM-based modeling. To preserve the most effective modeling paradigm for each modality, \name employs continuous diffusion for numerical columns and language models for categorical and free-form textual columns. However, rather than relying on autoregressive language models, we adopt Masked Diffusion Language Models (MDLMs) as the backbone architecture. MDLMs offer two key advantages. First, their diffusion-based formulation enables \name to model numerical and language modalities within a unified generative process. Second, unlike autoregressive models, MDLMs employ bidirectional attention, enabling the model to capture mutual dependencies among tabular columns. To enable joint numerical–language diffusion within a single MDLM, we introduce a trainable numerical tokenization module into the MDLM architecture, which allows continuous diffusion to represent each numerical value with a single token. As a result, \name learns to \textbf{jointly denoise numerical values and language content} during training and generates synthetic tabular data by simultaneously performing continuous diffusion and masked language diffusion during inference. The overview of \name is shown in Figure~\ref{fig:overview}.
We evaluate \name across a range of scenarios and benchmarks. On tables with free-form text, which specialized tabular diffusion models cannot generate, \name outperforms all language-model baselines. On conventional structured tables, it outperforms language-model baselines and remains competitive with specialized tabular diffusion models.

%% file: 02-methods.tex
\section{Methods}
\label{sec:methods}
\begin{figure*}[t]

\includegraphics[width=1.0\textwidth]{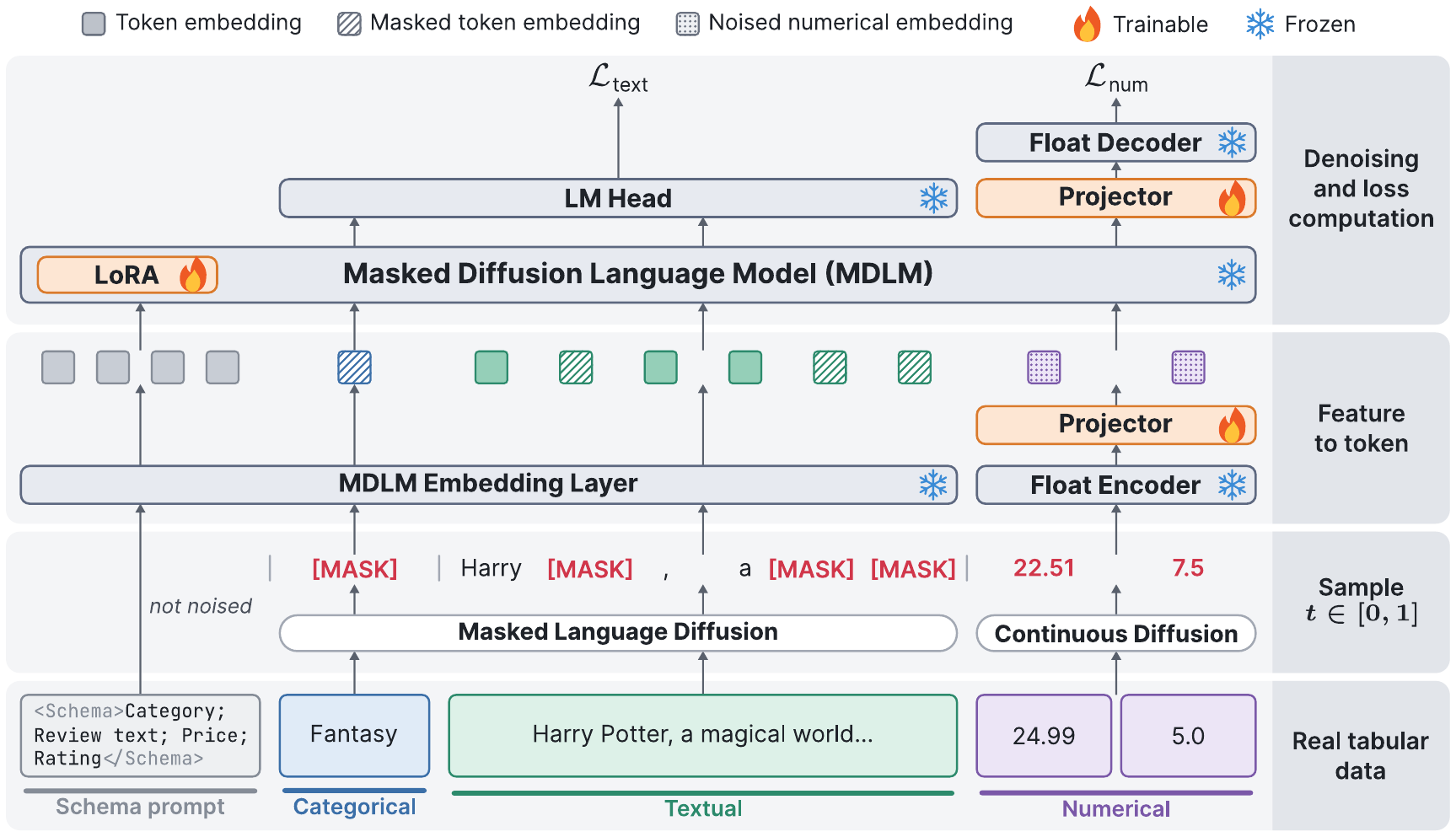}
\caption{Overview of \name. During training, given an input tabular sample, \name applies masked language diffusion to categorical and free-form textual features, and continuous diffusion to numerical features. Noisy inputs are mapped to token embeddings via a textual embedding layer and a numerical encoder. An MDLM then denoises these embeddings to reconstruct the original sample. Training updates only the projector in the numerical encoder and decoder, as well as the LoRA modules within each MDLM layer.
}
\label{fig:overview}
\end{figure*}

\subsection{Preliminary}
\paragraph{Problem setup and notation.}
Let $\mathcal{D}=\{ \mathbf{x}^{(i)} \}_{i=1}^{N}$ denote a tabular dataset with $M$ columns.
Each record is $\mathbf{x}^{(i)}=(x^{(i)}_{1},\dots,x^{(i)}_{M})$, where columns can be either numerical, categorical, or textual.
We use index sets $\mathcal{I}_{\text{num}}$, $\mathcal{I}_{\text{cat}}$, and $\mathcal{I}_{\text{text}}$ to denote numerical, categorical, and text columns, respectively.
Our goal is to learn a generative model $p_{\theta}(\mathbf{x})$ that can sample realistic records while capturing cross-column dependencies.

\paragraph{Continuous diffusion.}
We model the distribution of a continuous vector $\mathbf{z}_{0}\in\mathbb{R}^{d}$ using a diffusion process defined by the stochastic differential equation (SDE) $\mathrm{d}\mathbf{z} = \mathbf{f}(\mathbf{z},t)\mathrm{d}t + g(t)\mathrm{d}\mathbf{w}$.
In the variance-exploding (VE) formulation~\cite{ScoreSDE}, we set $\mathbf{f}(\mathbf{z},t)=\mathbf{0}$ and $g(t)=\sqrt{2\dot{\sigma}(t)\sigma(t)}$, where $\sigma(t):[0,1]\rightarrow \mathbb{R}_{+}$ is a strictly increasing function governing the noise level and $\dot{\sigma}(t)=\frac{\mathrm{d}\sigma(t)}{\mathrm{d}t}$ denotes its time derivative.
From a probabilistic perspective, this SDE corresponds to a continuous-time limit of a Markov chain where Gaussian noise is progressively added to the data.
The transition kernel $q(\mathbf{z}_{t}\mid \mathbf{z}_{s})$ for $s<t$ and the marginal $q(\mathbf{z}_{t}\mid \mathbf{z}_{0})$ are given by:
\begin{equation}
\begin{aligned}
\label{eq:numerical_forward}
q(\mathbf{z}_{t}\mid \mathbf{z}_{s}) &= \mathcal{N}\!\left(\mathbf{z}_{s},\, (\sigma^2(t)-\sigma^2(s))\mathbf{I}\right),\\
q(\mathbf{z}_{t}\mid \mathbf{z}_{0}) &= \mathcal{N}\!\left(\mathbf{z}_{0},\,\sigma^2(t)\mathbf{I}\right).
\end{aligned}
\end{equation}
The reverse generative process solves the probability flow ODE~\cite{ScoreSDE}, as follows:
\begin{equation}
\label{eq:numerical_backward}
\mathrm{d}\mathbf{z} = -\frac{1}{2}g(t)^2 \nabla_{\mathbf{z}}\log p_t(\mathbf{z}) \mathrm{d}t = -\dot{\sigma}(t)\sigma(t) \nabla_{\mathbf{z}}\log p_t(\mathbf{z}) \mathrm{d}t.
\end{equation}
We use a parameterized denoising model $\boldsymbol{\epsilon}_{\theta}(\mathbf{z}_{t},t)$ that estimates the score function via $\nabla_{\mathbf{z}}\log p_t(\mathbf{z}) \approx -\boldsymbol{\epsilon}_{\theta}(\mathbf{z}_{t},t)/\sigma(t)$.
Training minimizes the denoising error:
\begin{equation}
\mathcal{L}_{\mathrm{diff}}
=\mathbb{E}_{t,\mathbf{z}_{0},\boldsymbol{\epsilon}}
\left[\left\|\boldsymbol{\epsilon}-\boldsymbol{\epsilon}_{\theta}(\mathbf{z}_{0}+\sigma(t)\boldsymbol{\epsilon},t)\right\|_{2}^{2}\right],
\ \ \boldsymbol{\epsilon}\sim\mathcal{N}(\mathbf{0},\mathbf{I}).
\end{equation}

\paragraph{Masked diffusion language models (MDLMs).}
MDLMs~\citep{sahoo2024simple,llada} can be viewed as a discrete-state diffusion process~\cite{D3PM} over token sequences.
Let $\mathbf{s}_{0}=(s_{1},\dots,s_{L})$ be a length-$L$ sequence with tokens in a vocabulary $\mathcal{V}$ augmented with an absorbing mask state $m=[\mathrm{MASK}]$.
Using a token-wise Markov chain, the forward noising kernel factorizes as
\begin{equation}
q(\mathbf{s}_{t}\mid \mathbf{s}_{t-1})=\prod_{j=1}^{L} q(s_{t,j}\mid s_{t-1,j}),
\end{equation}
with a masking schedule $\beta_{t}\in[0,1]$ and absorbing transitions
\begin{equation}
\label{eq:category_forward}
q(s_{t,j}\mid s_{t-1,j})=
\begin{cases}
1, & s_{t-1,j}=m,\ s_{t,j}=m,\\
\beta_{t}, & s_{t-1,j}\neq m,\ s_{t,j}=m,\\
1-\beta_{t}, & s_{t-1,j}\neq m,\ s_{t,j}=s_{t-1,j},\\
0, & \text{otherwise},
\end{cases}
\end{equation}
Equivalently, the marginal has the closed form $q(s_{t,j}=s_{0,j}\mid s_{0,j})=\bar{\alpha}_{t}$ and $q(s_{t,j}=m\mid s_{0,j})=1-\bar{\alpha}_{t}$, where $\bar{\alpha}_{t}=\prod_{\tau=1}^{t}(1-\beta_{\tau})$.
The reverse model uses a bidirectional Transformer to predict the original token distribution from the partially masked sequence:
\begin{equation}
\label{eq:category_backward}
p_{\theta}(\mathbf{s}_{0}\mid \mathbf{s}_{t})
=\prod_{j=1}^{L} p_{\theta}(s_{0,j}\mid \mathbf{s}_{t}),
\end{equation}
and is typically trained by maximizing the log-likelihood of the ground-truth tokens at corrupted (masked) positions:
\begin{equation}
\mathcal{L}_{\mathrm{MDLM}}
=\mathbb{E}_{t,\mathbf{s}_{0},\mathbf{s}_{t}}
\left[-\sum_{\{j:\,s_{t,j}=[\text{MASK}]\}}\log p_{\theta}(s_{0,j}\mid \mathbf{s}_{t})\right].
\end{equation}
At sampling time, MDLMs start from an all-mask sequence and iteratively unmask tokens according to $p_{\theta}$.

\subsection{The Model Architecture of \name}\label{sec:model_arc}
We first present the architecture of \name. As illustrated in Figure~\ref{fig:overview}, \name comprises five main components: (1) a numerical encoder module, (2) a text embedding layer, (3) a masked diffusion language model, (4) a numerical decoder module, and (5) an LM head. We describe each component in detail below. Note that in this section we use $\hat{x}^{(i)}_j$ instead of $x^{(i)}_j$ to denote the noisy version of the input. We will describe how we obtain the $\hat{x}^{(i)}_j$ from $x^{(i)}_j$ in Section~\ref{sec:foward}.

\paragraph{Number encoder module.} Given an input numerical value $\hat{x}^{(i)}_{j}\in \mathbb{R}, j \in \mathcal{I}_{\text{num}}$, the number encoder module transforms the number into a token embedding with the same size as the MDLM. To achieve this, we first use quantile normalization~\cite{bolstad2003comparison} to standardize the numerical values as done in previous works~\cite{griffin, tabdiff}. Then, we use a pretrained Multi-Layer Perceptron (MLP) as the float number encoder to convert the normalized values into an $r$-dimensional embedding vector. Finally, a trainable projection is used to align the $r$-dimensional embedding vector into the $d$-dimensional MDLM embedding space:
\begin{equation}\label{eq:8}
 \mathbf{\hat{z}}^{(i)}_{j} = \text{ENC}(\hat{x}^{(i)}_{j}),\ \ \mathbf{z}^{(i)}_{j} = \text{PROJ}_{e}(\mathbf{\hat{z}}^{(i)}_{j}), \ \  j \in \mathcal{I}_{\text{num}},
\end{equation}
where $\text{ENC}$ and $\text{PROJ}_{e}$ denote the MLP-based encoder and projection module, respectively, with $\hat{\mathbf{z}}^{(i)}_{j} \in \mathbb{R}^{r}$ and $\mathbf{z}^{(i)}_{j} \in \mathbb{R}^{d}$. In our implementation, we pretrained $\text{ENC}$ based on Griffin~\cite{griffin} and keep it fixed throughout all experiments, while $\text{PROJ}_{e}$ is jointly optimized during training.

\paragraph{Text embedding layer.} The text embedding layer transforms textual input features $\hat{x}^{(i)}_{j}, j \in \mathcal{I}_{\text{cat}} \cup \mathcal{I}_{\text{text}}$ into a token embedding sequence. We directly adopt the pretrained embedding table from the underlying MDLM. Formally,
\begin{equation}\label{eq:9}
\mathbf{z}^{(i)}_j=(\mathbf{z}^{(i)}_{j, 1}, \ldots, \mathbf{z}^{(i)}_{j, l^i_j}) = \text{EMB}(\hat{x}^{(i)}_{j}), j \in \mathcal{I}_{\text{cat}} \cup \mathcal{I}_{\text{text}},
\end{equation}
where each token embedding $\mathbf{z}^{(i)}_{j, k} \in \mathbb{R}^{d}, k\in [l^{i}_j]$ and $l^{i}_j$ denotes the token sequence length of column $j$ in sample $i$, as each textual column may be mapped to multiple tokens in the MDLM embedding space.

\paragraph{Masked diffusion language model.} The masked diffusion language model is employed to jointly denoise both numerical and textual tokens. Formally, given an input sequence of token embeddings $\mathbf{z}^{(i)} =(\mathbf{z}^{(i)}_1, \ldots, \mathbf{z}^{(i)}_S)$ of length $S$ and a diffusion time step $t$, the MDLM outputs a denoised sequence $\mathbf{o}^{(i)}$ that estimates the corresponding representations at time $t=0$. In \name, the input token sequence is constructed by concatenating four components: a schema prompt, categorical features, textual features, and numerical features. The schema prompt provides a textual description of the feature type and the semantic meaning of each column, which is fixed for a given dataset. In detail, the input token sequence for sample $i$ is represented as:
\begin{equation}\label{eq:input_seq}
\begin{aligned}
\mathbf{z}^{(i)} = \big(\mathbf{z}_{p} , ( \mathbf{z}_j^{(i)}|j \in \mathcal{I}_{\text{cat}} \cup \mathcal{I}_{\text{text}}),(\mathbf{z}_j^{(i)} |j \in \mathcal{I}_{\text{num}} )\big),
\end{aligned}
\end{equation}
where $\mathbf{z}_p$ is the token embedding sequence for the schema prompt and $\mathbf{z}_j^{(i)} = (\mathbf{z}^{(i)}_{j, 1}, \ldots, \mathbf{z}^{(i)}_{j, l^i_j})$ for any $j \in \mathcal{I}_{\text{cat}} \cup \mathcal{I}_{\text{text}}$. 
The forward process of MDLM can be described as:
\begin{equation}\label{eq:10}
\mathbf{o}^{(i)} = \text{MDLM}(\mathbf{z}^{(i)}, t).
\end{equation}

For the MDLM architecture, we adopt a standard transformer architecture with bidirectional attention~\cite{transformer, bert}. For numerical tokens, an additional positional embedding is added to encode the diffusion noise level applied to the input, following the design in DiT~\cite{dit}. No such embedding is added for textual tokens, as the noise level can be implicitly captured by the proportion of masked tokens.

\paragraph{Number decoder module.} The number decoder module is used to decode the output token embedding for the numerical value back to the original number. We use a symmetric version of the number encoder module with one trainable projection, along with a pretrained MLP decoder:
\begin{equation}\label{eq:11}
 \mathbf{\bar{o}}^{(i)}_{j} = \text{PROJ}_{d}(\mathbf{o}^{(i)}_{j}),\ \ \bar{x}^{(i)}_{j} = \text{DEC}( \mathbf{\bar{o}}^{(i)}_{j}), \ \  j \in \mathcal{I}_{\text{num}},
\end{equation}
where $\text{DEC}$ and $\text{PROJ}_{d}$ denote the MLP-based decoder and projection module, respectively, with $\mathbf{o}^{(i)}_{j} \in \mathbb{R}^{d}$ representing the output token embedding from the MDLM and $\bar{\mathbf{o}}^{(i)}_{j} \in \mathbb{R}^{r}$. Similarly, the $\text{DEC}$ is pretrained jointly with the float number encoder and fixed during the experiment. 

\paragraph{LM head.} Finally, the LM head is used to decode the output token embedding for the textual value back to the original text. We directly use the frozen LM head from MDLM:
\begin{equation}\label{eq:12}
\bar{x}^{(i)}_{j} = \text{HEAD}(\mathbf{o}^{(i)}_j), \quad j \in \mathcal{I}_{\text{cat}} \cup \mathcal{I}_{\text{text}}.
\end{equation}

\subsection{The Forward Process of \name}
\label{sec:foward}
The forward diffusion process typically adds noise to the input. In \name, this process is applied to both numerical and textual modalities; for clarity, we describe the forward process for each modality separately.

\paragraph{Forward process of numerical features.} For numerical features, we adopt continuous diffusion to better capture fine-grained distributions. Given a clean input value $x^{(i)}_{j}, j \in \mathcal{I}_{\text{num}}$, we first sample a time $t \in [0, 1]$. Then, according to Equation~\ref{eq:numerical_forward}, the forward process adds noise as follows:
\begin{equation}
\hat{x}^{(i)}_{j} = x^{(i)}_{j} + \sigma(t)\epsilon, \ \  \epsilon\sim\mathcal{N}(0,1), \ \ j \in \mathcal{I}_{\text{num}}.
\end{equation}
Note that the noise is added to the normalized value.

\paragraph{Forward process of textual features.} For textual features, we adopt discrete masked diffusion, as used in MDLMs. Given a text sequence $x^{(i)}_j=(x^{(i)}_{j,1}, \ldots, x^{(i)}_{j, l^{i}_j}), j \in \mathcal{I}_{\text{cat}} \cup \mathcal{I}_{\text{text}}$, we first sample a time $t \in [0, 1]$. Then, according to Equation~\ref{eq:category_forward} and the accompanying discussion, the forward process independently transforms each token into a mask token with probability $1-\bar{\alpha}_{t}$, resulting in the masked token sequence $\hat{x}^{(i)}_{j}=(\hat{x}^{(i)}_{j,1}, \ldots, \hat{x}^{(i)}_{j, l^{i}_j})$.

Finally, in \name, we use a single $t$ for both continuous diffusion and discrete masked diffusion. This design enables the model to jointly denoise numerical and textual modalities under a unified noise level.

\subsection{Training of \name}
During training, we optimize \name to jointly denoise both the numerical modality and textual modality. Specifically, given an output sequence $\mathbf{o}^{(i)} = \big(\mathbf{o}_{p}, (\mathbf{o}_j^{(i)}|j \in \mathcal{I}_{\text{cat}} \cup \mathcal{I}_{\text{text}}), (\mathbf{o}_j^{(i)} |j \in \mathcal{I}_{\text{num}} )\big)$ from MDLM, we first transform the numerical token embedding back to real numbers using the number decoder module and then the loss is computed separately for continuous diffusion and masked diffusion:
\begin{equation}
\mathcal{L}_{\text{num}} = \frac{1}{N}\sum_{i=1}^{N}\sum_{j\in \mathcal{I}_{\text{num}}}\left\|x^{(i)}_{j}-\bar{x}^{(i)}_{j}\right\|_{2}^{2}, \quad j\in \mathcal{I}_{\text{num}},
\end{equation}
\begin{equation}
\begin{aligned}
\mathcal{L}_{\text{text}} = &\frac{1}{N}\sum^N_{i=1}\sum_{\hat{x}^{(i)}_{j, k}=[\text{MASK}]}\text{CE} <\bar{x}^{(i)}_{j,k}, x^{(i)}_{j,k}>, \\ & k\in[l^i_j], \quad j \in \mathcal{I}_{\text{cat}} \cup \mathcal{I}_{\text{text}},
\end{aligned}
\end{equation}

where $\text{CE}<\cdot,\cdot>$ is the cross entropy loss. The final objective is optimized using a step-dependent weighting schedule:
\begin{equation}
\mathcal{L}=\mathcal{L}_{\text{text}}+\lambda(s)\,\mathcal{L}_{\text{num}},\ \ 
\lambda(s)=\lambda_{\max}\cdot \min\left(1,\frac{s}{s_{\text{warm}}}\right),
\end{equation}
where $s$ denotes the global optimization step, and this warm-up schedule stabilizes the alignment of newly initialized numerical projections with the pretrained MDLM backbone. We use default hyperparameters $\lambda_{\max}=1$ and $s_{\text{warm}}=2000$.
In practice, we freeze the pretrained weights of the MDLM and introduce trainable LoRA modules~\cite{lora} into both the feed-forward and attention components of each transformer layer.

\subsection{The backward sampling in \name}
Finally, we describe the sampling process of \name. After training, generation starts from noise and progressively denoises the inputs to produce synthetic tabular samples following the learned data distribution. For numerical modality, the process is initialized with pure Gaussian noise, while for textual modality, it starts from a fully masked token sequence. We then iteratively apply Equation~\ref{eq:numerical_backward} and Equation~\ref{eq:category_backward} to the numerical and textual modalities to recover clean data. More details are provided in Appendix~\ref{apdx:impl}.

%% file: 03-relatedwork.tex
\section{Related Works}
Existing methods in tabular data generation can be classified into three categories based on their underlying frameworks.

\paragraph{VAE/GAN-based tabular generators.} This line of work formulates tabular data generation using Variational Autoencoders (VAEs)~\citep{kingma2013auto} or Generative Adversarial Networks (GANs)~\citep{goodfellow2014generative}. Representative methods include CTGAN and TVAE~\citep{xu2019modeling}. GOGGLE~\citep{liu2023goggle} further enhances this paradigm by explicitly modeling column dependencies through a Graph Neural Network–augmented VAE architecture. However, these approaches often lack sufficient expressivity when confronted with complex tabular distributions involving intricate feature interactions.

\paragraph{Diffusion-based tabular generators.}
Motivated by the strong generative capacity of diffusion models~\citep{DDPM}, a growing body of work adapts diffusion processes for tabular data generation. Early methods model numerical and categorical features using separate discrete-time diffusion~\citep{D3PM}, as in~\citet{tabddpm,lee2023codi}, but discretization can lead to looser ELBO bounds and suboptimal generation quality~\citep{ScoreSDE,kingma2021variational}. More recent approaches encode tabular features into continuous latent spaces and apply Gaussian diffusion~\citep{zheng2022diffusion,zhang2024mixedtype}. However, such latent modeling introduces additional encoding overhead and may only indirectly capture heterogeneous feature interactions. Recently, CDTD~\citep{mueller2023continuous} explores feature-wise noise schedules under continuous diffusion, and TabDiff~\citep{tabdiff} further extends it to mixed-type feature-level diffusion.
Diffuse Everything~\citep{rojas2025diffuse} couples continuous and masked discrete diffusion through decoupled per-modality noise times and is applied to text--image generation and mixed-type tabular data generation; its tabular instantiation draws every categorical column from a fixed category set, with no free-form text column. TabNAT~\citep{tabnat}, designed specifically for tables, pairs continuous diffusion for numerical features with a bidirectional masked Transformer over column-specific categorical indices, so its discrete generation is likewise restricted to fixed category sets.
In contrast, \name leverages a pretrained MDLM with a large semantic vocabulary to jointly generate numerical, categorical, and free-form textual fields within a single model.

\paragraph{Language model-based tabular generators.}
Recent advances in large language models have also inspired language model–based approaches for tabular data generation. These methods serialize each row into a text sequence and fine-tune an autoregressive language model to capture row-level distributions, as exemplified by GReaT~\citep{GreaT} with a GPT-2~\citep{radford2019language} backbone. 
DiffLM~\citep{difflm} integrates VAEs and latent diffusion within a language modeling framework. However, DiffLM decodes each row, numbers included, with a frozen autoregressive LLM, so numerical values are generated as digit tokens. TabMT~\citep{gulati2023tabmt} and TabularMDLM~\citep{isomura2025tabularmdlm} generate rows by masked token prediction, but likewise represent every value, numbers included, as discrete tokens. To the best of our knowledge, \name is the first method for tabular data in which one pretrained masked diffusion language model denoises numerical values by continuous diffusion and categorical values and free-form text by token-level masked diffusion.

%% file: 04-experiments.tex
\section{Experiments}
In this section, we conduct extensive experiments to evaluate the proposed \name. Specifically, we aim to answer the following questions. \textbf{Q1:} Can \name effectively model the mutual dependencies among numerical, categorical, and text columns? \textbf{Q2:} How well does \name perform on real-world tabular data generation tasks involving free-form text columns? \textbf{Q3:} Can \name achieve performance on par with existing methods on tasks that do not include textual columns? We implement \name based on LLaDA-8B~\cite{llada} and \names based on MDLM~\citep{sahoo2024simple} (170M parameters) to examine whether our framework remains effective with a much smaller backbone. Other implementation details can be found in Appendix~\ref{apdx:impl}. 

\subsection{Results on synthetic tabular datasets}
\label{sec:eval_syn}
In this section, we answer the question \textbf{Q1} through two carefully designed synthetic tabular datasets.

\paragraph{Datasets.} Existing real-world tabular generation datasets predominantly focus on numerical and categorical features and lack text fields. Thus, we construct two synthetic tabular datasets whose text column is rendered from the numerical and categorical columns: \textbf{MathExpr} and \textbf{ProfileBio}. Each MathExpr sample includes two floating-point variables, categorical features indicating the unary and binary operators applied to them, and a textual column containing the corresponding LaTeX expression. ProfileBio contains numerical and categorical attributes describing a person’s demographic and educational background, along with a textual biography generated from the other columns. For both datasets, accurate generation requires models to \textbf{capture correlations between numerical, categorical, and textual columns}. Examples are shown in Tables~\ref{tab:mathexpr_example} and~\ref{tab:profilebio_example}, and the generation process in Appendix~\ref{apdx:dataset}. Both datasets serve as controlled diagnostics: the text field is rendered from the structured fields by a rule never shown to the models, so the consistency between the text and the structured fields can be checked exactly for every generated row, which real free-form text does not allow. Open-ended text generation is evaluated on the real-world datasets in Section~\ref{sec:eval_rel_text}.

\paragraph{Baselines.} Given these two datasets contain text columns, which specialized tabular diffusion models cannot generate, we compare the proposed \name with autoregressive and masked diffusion LLMs. Specifically, we include Qwen2.5 (7B/14B) with 3-shot in-context learning (ICL), and Qwen2.5-7B and LLaDA-8B with LoRA-based supervised fine-tuning (SFT). We further include GReaT~\cite{GreaT}, which fully fine-tunes GPT-2, and DiffLM~\cite{difflm}, which trains a latent diffusion model on top of a VAE with a frozen Mistral-7B~\citep{jiang2023mistral7b} decoder.

\paragraph{Metrics.} We evaluate generated data along two groups. \textbf{1) Fidelity:} Shape and Trend measure marginal column-wise similarity and pairwise dependencies among numerical and categorical columns, both reported as error rates; \textbf{2) Cross-field consistency:} for MathExpr, we report Operation Match Rate (Op-MR) and Expression Match Rate (Exp-MR), measuring whether operators and numerical literals in the generated LaTeX align with the structured fields. For ProfileBio, we report Biography Match Rate (Bio-MR), checking consistency between the generated biography and structured attributes. Detailed definitions are provided in Appendix~\ref{apdx:metrics}.

\begin{table}[h]
\vspace{-10pt}
\centering
\footnotesize
\setlength{\tabcolsep}{6.0pt}
\caption{Evaluation results on the \textbf{MathExpr} and \textbf{ProfileBio} datasets (\%).}
\label{tab:mathexpr_profilebio}
\begin{tabular}{lccccccc}
\toprule
\multirow{2}{*}{\textbf{Method}}
& \multicolumn{4}{c}{\textbf{MathExpr}}
& \multicolumn{3}{c}{\textbf{ProfileBio}} \\
\cmidrule(lr){2-5} \cmidrule(lr){6-8}
& Shape $\downarrow$
& Trend $\downarrow$
& Op-MR $\uparrow$
& Exp-MR $\uparrow$
& Shape $\downarrow$
& Trend $\downarrow$
& Bio-MR $\uparrow$ \\
\midrule
GReaT
& $19.28$ & $25.40$ & $97.67$ & $93.07$
& \redbf{$\mathbf{2.76}$} & $10.90$ & $97.89$ \\
DiffLM
& $6.84$ & $17.70$ & $97.31$ & $97.12$
& $18.82$ & $31.18$ & $68.74$ \\
Qwen2.5-7B$_{\text{ICL}}$
& $39.04$ & $56.65$ & $53.93$ & $53.92$
& $41.29$ & $64.52$ & $7.97$ \\
Qwen2.5-14B$_{\text{ICL}}$
& $25.61$ & $36.07$ & $76.01$ & $75.12$
& $43.97$ & $65.00$ & $13.00$ \\
Qwen2.5-7B$_{\text{SFT}}$
& $5.07$ & $50.81$ & $99.68$ & $99.34$
& $10.51$ & $36.26$ & $98.51$ \\
LLaDA-8B$_{\text{SFT}}$
& $5.41$ & $42.54$ & $98.36$ & $97.90$
& $5.81$ & $44.14$ & $97.63$ \\
\midrule
\textbf{\names}
& \redbf{$\mathbf{2.12}$} & \redbf{$\mathbf{4.42}$} & $99.00$ & $96.92$
& $5.90$ & $9.41$ & $86.59$ \\
\textbf{\name}
& $3.69$ & $6.07$ & \redbf{$\mathbf{99.99}$} & \redbf{$\mathbf{99.80}$}
& $4.84$ & \redbf{$\mathbf{7.81}$} & \redbf{$\mathbf{98.60}$} \\
\bottomrule
\end{tabular}
\vspace{-10pt}
\end{table}

\paragraph{Results.} As shown in Table~\ref{tab:mathexpr_profilebio}, our method achieves the lowest Trend error and the highest match rates on both MathExpr and ProfileBio. On MathExpr, both variants of our model obtain lower Shape and Trend errors than every baseline: compared with the strongest baseline on each metric, \name reduces them by $27.2\%$ and $65.7\%$. Since MathExpr requires the generated LaTeX expression to be consistent with both numerical values and categorical operator columns, these results indicate that \name can preserve explicit dependencies across numerical, categorical, and textual fields rather than merely matching marginal distributions. On ProfileBio, the highest Match Rate shows that \name generates biographies that remain semantically aligned with the corresponding structured attributes. Together, the strong fidelity and cross-field consistency results provide a direct answer to \textbf{Q1}: \name can effectively model mutual dependencies among heterogeneous tabular fields through joint numerical-language denoising. \names recovers the structured dependencies of both datasets, whereas its $12.0$-point lower Bio-MR on ProfileBio suggests that the 8B backbone helps keep the generated text consistent with the structured fields.

\subsection{Results on real-world tabular datasets with free-form text features}
\label{sec:eval_rel_text}
In this section, we evaluate \name on real-world tabular datasets with free-text to answer \textbf{Q2}.

\paragraph{Dataset and Baselines.} We use two real-world datasets, Amazon and Arxiv, both derived from the RelBench~\cite{robinson2024relbench} relational benchmark. Amazon merges several \texttt{rel-amazon} tables into one table with numerical, categorical, and long-text fields (e.g., title, description, and review); Arxiv is a paper-level table from \texttt{rel-arxiv} with numerical, categorical, and free-form text fields (e.g., title, arxiv code, and abstract). Details are in Appendix~\ref{apdx:dataset}, and the baselines are the same as those in Section~\ref{sec:eval_syn}.

\begin{table}[h]
\centering
\footnotesize
\setlength{\tabcolsep}{3.0pt}
\vspace{-10pt}
\caption{Results on the \textbf{Amazon} and \textbf{Arxiv} datasets. ($^*$ indicates \textit{(w/o nums)})}
\label{tab:amazon_arxiv}
\begin{tabular}{lcccccccc}
\toprule
\multirow{2}{*}{\textbf{Method}}
& \multicolumn{4}{c}{\textbf{Amazon}}
& \multicolumn{4}{c}{\textbf{Arxiv}} \\
\cmidrule(lr){2-5} \cmidrule(lr){6-9}
& Shape (\%)$\downarrow$
& Trend (\%)$\downarrow$
& $\text{MLE}^*$ $\uparrow$
& MLE $\uparrow$
& Shape (\%)$\downarrow$
& Trend (\%)$\downarrow$
& $\text{MLE}^*$ $\uparrow$
& MLE $\uparrow$ \\
\midrule
Real
& $0.0$ & $0.0$ & $.905$ & $.906$
& $0.0$ & $0.0$ & $.968$ & $.968$ \\
\midrule
GReaT
& $6.91$ & $5.94$ & $.874$ & $.877$
& $6.72$ & $6.43$ & $.599$ & $.603$ \\
DiffLM
& $17.23$ & $43.84$ & $.819$ & $.817$
& $15.17$ & $16.38$ & $.848$ & $.837$ \\
Qwen2.5-7B$_{\text{ICL}}$
& $37.93$ & $68.27$ & $.636$ & $.629$
& $68.04$ & $42.34$ & $.521$ & $.516$ \\
Qwen2.5-14B$_{\text{ICL}}$
& $39.19$ & $78.22$ & $.684$ & $.679$
& $40.79$ & $34.35$ & $.556$ & $.527$ \\
Qwen2.5-7B$_{\text{SFT}}$
& $11.41$ & $46.31$ & $.824$ & $.834$
& $11.19$ & $10.02$ & $.701$ & $.727$ \\
LLaDA-8B$_{\text{SFT}}$
& $7.76$ & $37.13$ & $.891$ & $.892$
& $14.10$ & $43.81$ & $.931$ & $.930$ \\
\midrule
\textbf{\names}
& $9.15$ & $7.48$ & $.861$ & $.864$
& $10.15$ & $8.05$ & $.944$ & \redbf{$\mathbf{.951}$} \\
\textbf{\name}
& \redbf{$\mathbf{4.67}$} & \redbf{$\mathbf{5.33}$} & \redbf{$\mathbf{.893}$} & \redbf{$\mathbf{.895}$}
& \redbf{$\mathbf{6.30}$} & \redbf{$\mathbf{6.30}$} & \redbf{$\mathbf{.946}$} & $.948$ \\
\bottomrule
\vspace{-20pt}
\end{tabular}
\end{table}

\paragraph{Metrics.} Similar to Section~\ref{sec:eval_syn}, we evaluate \textbf{1) Fidelity:} Shape and Trend. Additionally, we include \textbf{2) Downstream utility:} Machine Learning Efficiency (MLE), 
which measures how well predictive models trained on synthetic data generalize to real test data. Finally, to evaluate cross-field consistency between free-form text and the remaining numerical and categorical features, we follow an MLE-like protocol: we first serialize all non-numerical fields into structured text and encode them with the sentence embedding model Nomic~\cite{nussbaum2025nomic}, then train an XGBoost Classifier~\cite{Chen2016XGBoostAS} using either (i) text embeddings only or (ii) text embeddings concatenated with numerical features, which isolates the contribution of generated text/categorical fields. MLE is reported as AUC.

\paragraph{Results.}
The results on Amazon and Arxiv are presented in Table~\ref{tab:amazon_arxiv}. Our method achieves the best performance across Shape, Trend, and MLE on both datasets. Compared with LLaDA-8B$_{\text{SFT}}$, which shares the backbone and the training data of \name but represents numerical values as digit tokens, \name reduces the Shape and Trend errors by $39.8$--$85.6\%$ on both tables. GReaT is the strongest baseline on Shape and Trend but has a substantially lower downstream utility on Arxiv. These results show that \name effectively preserves structured-field marginal distributions and pairwise structured dependencies in real-world datasets that also contain free-form textual fields. The distribution of every numerical column is compared with the real data in Appendix~\ref{apdx:marginals}.

\begin{figure}[t]
\centering
\includegraphics[width=\linewidth]{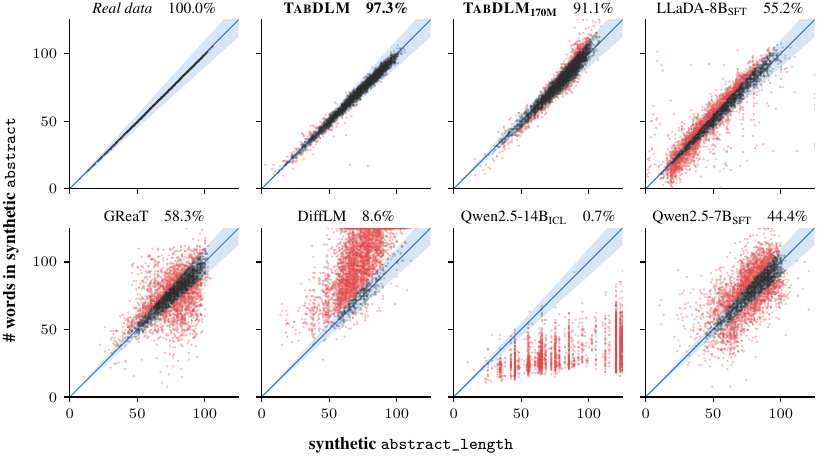}
\caption{Number--text consistency on Arxiv. Each point is a synthetic row: its \texttt{abstract\_length} ($x$) against the word count of its \texttt{abstract} ($y$). Each panel reports the percentage of rows whose word count is within $\pm10\%$ of \texttt{abstract\_length} (shaded band); red points fall outside the band.}
\label{fig:arxiv_agree}
\end{figure}

Regarding downstream utility, the MLE of our method closely approaches the real-data upper bound. This indicates that the generated text, categorical, and numerical fields jointly preserve label-relevant information. Using only the generated text and categorical fields (MLE$^*$), \name attains the highest score on both tables, within $.012$ and $.022$ of real data, so the generated text itself carries the label-relevant information. To test whether the generated numbers agree with the generated text, we compare the word count of each generated Arxiv \texttt{abstract} with the generated \texttt{abstract\_length} of its row (Figure~\ref{fig:arxiv_agree}): the share of rows within $\pm10\%$ is $39.0$ percentage points higher for \name than for the strongest baseline, so the numbers and text generated by \name are consistent with each other. We evaluate the generated text and present a case study in Appendices~\ref{apdx:textq} and~\ref{apdx:examples}.

Comparing our two variants, \name reduces the Shape and Trend errors of \names by $21.7$--$49.0\%$ and, on Amazon, generates text with higher judged consistency and fewer hallucinations (Appendix~\ref{apdx:textq}), so a stronger backbone brings further gains. \names nevertheless remains strong overall despite its much smaller backbone. Together, these results provide a direct answer to \textbf{Q2}: \name performs strongly on real-world tabular generation tasks involving free-form text, preserving both structured-field fidelity and downstream task-relevant information across modalities.

\subsection{Results on real-world tabular datasets without free-form text features}
\label{sec:eval_rel_notext}
Finally, we answer \textbf{Q3} on traditional tabular data generation benchmarks without free-form text features, enabling comparison with existing specialized methods that cannot generate such text.

\paragraph{Datasets, baselines, and metrics.} We conduct experiments on five real-world tabular datasets: Adult, Default, Shoppers, Magic, and Beijing. Detailed dataset profiles are presented in Appendix~\ref{apdx:real-tabular}. Baselines are widely-used tabular generators from four categories: 1) GAN-based: CTGAN~\cite{xu2019modeling};
2) VAE-based: TVAE~\cite{xu2019modeling} and GOGGLE~\cite{liu2023goggle};
3) Autoregressive Tabular Language Model: GReaT~\cite{GreaT} and DiffLM~\cite{difflm};
4) Diffusion-based: STaSy~\cite{kimstasy}, CoDi~\cite{lee2023codi}, TabDDPM~\cite{tabddpm}, TabSyn~\cite{zhang2024mixedtype}, TabDiff~\cite{tabdiff}, and TabNAT~\cite{tabnat}.
We evaluate along three groups of metrics, defined in Appendix~\ref{apdx:metrics}: \textbf{1) Fidelity:} Shape, Trend, $\alpha$-Precision, and C2ST; \textbf{2) Downstream utility:} Machine Learning Efficiency (MLE); \textbf{3) Privacy:} Distance to Closest Record (DCR) evaluates privacy risk by measuring the distance between each synthetic sample and its nearest training sample, where overly small distances may indicate memorization. We report Shape, Trend in the main paper and include $\alpha$-Precision, C2ST, MLE, and DCR results in Appendix~\ref{apdx:add_exps}.

\begin{table*}[htbp]
\centering
\footnotesize
\setlength{\tabcolsep}{5pt}
\caption{Results on the error rates (\%) of \textbf{Shape}~$\downarrow$~/~\textbf{Trend}~$\downarrow$. Each cell reports Shape\,/\,Trend. Full per-metric results are provided in Appendix~\ref{apdx:full_shape_trend}. (\redbf{B}: best overall; \textbf{B}: best in language-based model)}
\vspace{-5pt}
\label{tab:shape_trend}
\begin{tabular}{lccccc|>{\columncolor{gray!15}}c}
\toprule
\textbf{Method} & \textbf{Adult} & \textbf{Default} & \textbf{Shoppers} & \textbf{Magic} & \textbf{Beijing} & \textbf{Average} \\
\midrule
CTGAN & $16.84$\,/\,$20.23$ & $16.83$\,/\,$26.95$ & $21.15$\,/\,$13.08$ & $9.81$\,/\,$7.00$ & $21.39$\,/\,$22.95$ & $17.20$\,/\,$18.04$ \\
TVAE & $14.22$\,/\,$14.15$ & $10.17$\,/\,$19.50$ & $24.51$\,/\,$18.67$ & $8.25$\,/\,$5.82$ & $19.16$\,/\,$18.01$ & $15.26$\,/\,$15.23$ \\
GOGGLE & $16.97$\,/\,$45.29$ & $17.02$\,/\,$21.94$ & $22.33$\,/\,$23.90$ & $1.90$\,/\,$9.47$ & $16.93$\,/\,$45.94$ & $15.03$\,/\,$29.31$ \\
STaSy & $11.29$\,/\,$14.51$ & $5.77$\,/\,$5.96$ & $9.37$\,/\,$8.49$ & $6.29$\,/\,$6.61$ & $6.71$\,/\,$8.00$ & $7.89$\,/\,$8.71$ \\
CoDi & $21.38$\,/\,$22.49$ & $15.77$\,/\,$68.41$ & $31.84$\,/\,$17.78$ & $11.56$\,/\,$6.53$ & $16.94$\,/\,$7.07$ & $19.50$\,/\,$24.46$ \\
TabDDPM & $1.75$\,/\,$3.01$ & $1.57$\,/\,$4.89$ & $2.72$\,/\,$6.61$ & $1.01$\,/\,$1.70$ & $1.30$\,/\,$2.71$ & $1.67$\,/\,$3.78$ \\
TabSyn & $0.81$\,/\,$1.93$ & $1.01$\,/\,$2.81$ & $1.44$\,/\,$2.13$ & $1.03$\,/\,$0.88$ & $1.26$\,/\,$3.13$ & $1.11$\,/\,$2.18$ \\
TabDiff & \redbf{$\mathbf{0.63}$}\,/\,\redbf{$\mathbf{1.49}$} & $1.24$\,/\,$2.55$ & $1.28$\,/\,$1.74$ & $0.78$\,/\,\redbf{$\mathbf{0.76}$} & $1.03$\,/\,$2.59$ & $0.99$\,/\,\redbf{$\mathbf{1.83}$} \\
TabNAT & $0.73$\,/\,$1.61$ & \redbf{$\mathbf{0.80}$}\,/\,\redbf{$\mathbf{2.21}$} & \redbf{$\mathbf{1.11}$}\,/\,\redbf{$\mathbf{1.67}$} & \redbf{$\mathbf{0.62}$}\,/\,$1.37$ & \redbf{$\mathbf{0.79}$}\,/\,\redbf{$\mathbf{2.48}$} & \redbf{$\mathbf{0.81}$}\,/\,$1.87$ \\
\midrule
GReaT & $12.12$\,/\,$17.59$ & $19.94$\,/\,$70.02$ & $14.51$\,/\,$45.16$ & $16.16$\,/\,$10.23$ & $8.25$\,/\,$59.60$ & $14.20$\,/\,$40.52$ \\
DiffLM & $9.74$\,/\,-- & $9.06$\,/\,-- & $10.07$\,/\,-- & $7.53$\,/\,-- & $6.35$\,/\,-- & $8.55$\,/\,-- \\
Qwen2.5-7B$_{\text{ICL}}$ & $27.80$\,/\,$37.61$ & $22.39$\,/\,$31.02$ & $34.37$\,/\,$34.77$ & $13.41$\,/\,$20.62$ & $31.83$\,/\,$40.83$ & $25.96$\,/\,$32.97$ \\
Qwen2.5-14B$_{\text{ICL}}$ & $25.17$\,/\,$42.67$ & $21.52$\,/\,$28.72$ & $51.21$\,/\,$42.32$ & $17.51$\,/\,$18.11$ & $29.76$\,/\,$38.20$ & $29.03$\,/\,$34.00$ \\
Qwen2.5-7B$_{\text{SFT}}$ & $5.11$\,/\,$17.03$ & $2.58$\,/\,$16.99$ & $8.76$\,/\,$10.28$ & $7.53$\,/\,$14.45$ & $7.49$\,/\,$28.68$ & $6.29$\,/\,$17.49$ \\
LLaDA-8B$_{\text{SFT}}$ & $1.69$\,/\,$26.15$ & $2.00$\,/\,$26.30$ & $13.56$\,/\,$16.42$ & $5.60$\,/\,$13.71$ & $3.64$\,/\,$27.90$ & $5.30$\,/\,$22.10$ \\
\midrule
\textbf{\names} & $1.49$\,/\,$\mathbf{2.56}$ & $3.64$\,/\,$6.19$ & $4.54$\,/\,$4.62$ & $\mathbf{1.96}$\,/\,$\mathbf{1.76}$ & $\mathbf{2.39}$\,/\,$\mathbf{2.71}$ & $2.80$\,/\,$3.57$ \\
\textbf{\name} & $\mathbf{1.46}$\,/\,$2.74$ & $\mathbf{1.18}$\,/\,$\mathbf{2.33}$ & $\mathbf{2.05}$\,/\,$\mathbf{2.40}$ & $2.93$\,/\,$2.85$ & $2.42$\,/\,$2.93$ & $\mathbf{2.01}$\,/\,$\mathbf{2.65}$ \\
\bottomrule
\end{tabular}
\vspace{-13pt}
\end{table*}

\paragraph{Results.} As reported in Table~\ref{tab:shape_trend}, \name performs competitively with state-of-the-art tabular diffusion models on Shape and Trend. \textbf{This is notable because such tabular-specific synthetic methods operate in a relatively restricted feature space} (e.g., categorical columns are represented via one-hot vectors or limited discrete states), whereas \name is built to jointly model heterogeneous fields and ultimately support open-ended text generation, yet it still remains strong on purely numerical and categorical tabular datasets. \names also performs strongly with a $47\times$ smaller backbone, which offers an efficient option when no free text is involved (Appendix~\ref{apdx:cost}). Beyond this, \name consistently and substantially outperforms all language-based baselines, with average gains of $62.1\%$ in Shape and $84.8\%$ in Trend. 
Notably, \textbf{language-based generators often achieve reasonable Shape but markedly worse Trend}. A likely reason is that row serialization simplifies matching column-wise marginals, as each field can be generated locally to fit its token-level distribution. However, Trend depends on modeling \textbf{joint} dependencies among numerical and categorical columns, which is particularly difficult to preserve when numbers are serialized as multi-token strings.
In practice, numerical values are represented as subword token sequences with precision-sensitive semantics, whereas categorical fields act as discrete identifiers, making consistent numerical–categorical correlations hard to preserve under token-level generation. In contrast, \name jointly generates numerical and categorical features within a unified diffusion process and leverages bidirectional interactions across columns, enabling more direct modeling of tabular dependencies than token-serialized row generation and yielding greater fidelity on structured datasets.

\subsection{Ablation Study}\label{sec:ablation_study}
We assess the contribution of each component of \name with three variants: (i) \name-noFloatAE, which replaces the pretrained float encoder/decoder with randomly initialized counterparts; (ii) \name-onlyContDiff, which removes MDLM-based modeling for categorical fields; and (iii) \name-onlyMDLM, which removes the continuous diffusion branch and processes numerical features as serialized tokens. Detailed variant descriptions are provided in Appendix~\ref{apdx:ablation}.

\begin{table*}[htbp]
\centering
\footnotesize
\caption{Ablation study on the average performance across Adult, Default, and Shoppers.}
\vspace{-5pt}
\setlength{\tabcolsep}{6pt}
\label{tab:ablation}
\begin{tabular}{lccccc}
\toprule
\textbf{Method} & \textbf{Shape $\downarrow$} & \textbf{Trend $\downarrow$} & \textbf{MLE $\uparrow$} & \textbf{C2ST $\uparrow$} & \textbf{$\alpha$-Precision $\uparrow$} \\
\midrule
\name-{\text{noFloatAE}} & $2.15$ & $4.34$ & $.870$ & $0.9256$ & $98.50$ \\
\name-{\text{onlyContDiff}} & $4.19$ & $7.28$ & $.850$ & $0.8399$ & $91.70$ \\
\name-{\text{onlyMDLM}} & $5.75$ & $22.96$ & $.852$ & $0.7892$ & $88.58$ \\
\midrule
\textbf{\name} & \redbf{$\mathbf{1.56}$} & \redbf{$\mathbf{2.49}$} & \redbf{$\mathbf{.874}$} & \redbf{$\mathbf{0.9571}$}& \redbf{$\mathbf{98.56}$} \\
\bottomrule
\end{tabular}
\vspace{-5pt}
\end{table*}

The full \name consistently outperforms all three variants. The largest drop occurs in \name-onlyMDLM, supporting our motivation in Section~\ref{sec:intro} that subword tokenization fragments precision-sensitive numerical values and hampers fine-grained numerical–categorical correlation modeling. \name-noFloatAE and \name-onlyContDiff also degrade noticeably, validating the necessity of pretrained numerical encoding and MDLM-based modeling for categorical fields, respectively.

%% file: 06-conclusion.tex
\section{Conclusion}
In this work, we propose \name, to the best of our knowledge the first joint numerical--language diffusion method for generating tables with numerical, categorical, and free-form text columns. \name leverages an MDLM with special numerical tokenization to allow a single model to denoise numbers by continuous diffusion and categories and text by masked diffusion. Extensive experiments validate the effectiveness of \name over baseline models. Its generated numbers and text agree with each other, and its free-form text is both diverse and consistent with the other columns. We discuss limitations and future directions, including sampling efficiency and modality-specific noise schedules, in Appendix~\ref{sec:limitations}.

%% file: 08-appendix.tex
\appendix
\startcontents[appendix]
\section*{Contents of the Appendix}
\printcontents[appendix]{}{1}[2]{}
\clearpage

\section{Detailed Experiment Setups}\label{apdx: experiment}
\subsection{Datasets}\label{apdx:dataset}
\subsubsection{MathExpr}\label{apdx:mathexpr}
MathExpr is a synthetic dataset designed to evaluate the joint generation of heterogeneous tabular records involving numerical values, categorical operators, and LaTeX expressions rendered from the structured columns.
Each record contains two numerical columns, three categorical operator columns, and one text column:
\[
\mathbf{r} = (x_1, x_2, o_1, o_2, o_3, e_{\texttt{latex}}).
\]
Here, $o_1$ and $o_2$ are unary operators applied to $x_1$ and $x_2$, respectively, while $o_3$ is a binary operator that combines the two transformed terms. The text column $e_{\texttt{latex}}$ is a LaTeX expression deterministically constructed from the preceding structured columns.

We generate $10{,}000$ records and split them into a training/real set and a validation set with a ratio of $9{:}1$. During sampling, each model generates $9{,}000$ synthetic samples, and these synthetic samples are then used to compute the distribution fidelity metrics and expression consistency metrics described in Appendix~\ref{metric:syntab}.

\paragraph{Numerical value sampling.}
We sample $x_1$ and $x_2$ from discrete supports with step size $0.1$:
\[
x_1 \in \{0.1, 0.2, \ldots, 6.0\},
\qquad
x_2 \in \{3.0, 3.1, \ldots, 9.9\}.
\]
To induce a non-uniform yet diverse distribution, we sample $x_1$ and $x_2$ from an equal-weight mixture of a discrete Gaussian distribution and a uniform distribution over the corresponding support. The discrete Gaussian components are centered at $\mu_1=3$ and $\mu_2=6.5$, respectively, with shared standard deviation $\sigma=1.5$.

\paragraph{Categorical operator sampling.}
The unary operators $o_1$ and $o_2$ are sampled independently from
\[
\mathcal{O}_{\text{unary}}
=
\{
\texttt{none}, \texttt{log}, \texttt{exp}, \texttt{sqrt},
\texttt{sin}, \texttt{cos}, \texttt{tan},
\texttt{square}, \texttt{cube}
\}.
\]
Their categorical priors are:
\[
\begin{aligned}
p(o_1) = \{&
\texttt{none}:0.18,\ 
\texttt{log}:0.16,\ 
\texttt{sqrt}:0.13,\ 
\texttt{square}:0.12,\ 
\texttt{sin}:0.10,\\
&
\texttt{cos}:0.10,\ 
\texttt{tan}:0.07,\ 
\texttt{exp}:0.07,\ 
\texttt{cube}:0.07
\},\\[2mm]
p(o_2) = \{&
\texttt{none}:0.22,\ 
\texttt{sin}:0.14,\ 
\texttt{cos}:0.14,\ 
\texttt{sqrt}:0.12,\ 
\texttt{log}:0.10,\\
&
\texttt{square}:0.09,\ 
\texttt{tan}:0.07,\ 
\texttt{exp}:0.06,\ 
\texttt{cube}:0.06
\}.
\end{aligned}
\]
The binary operator $o_3$ is sampled from
\[
\mathcal{O}_{\text{binary}}
=
\{\texttt{add}, \texttt{sub}, \texttt{mul}, \texttt{div}\},
\]
with categorical prior
\[
p(o_3)
=
\{
\texttt{add}:0.35,\ 
\texttt{mul}:0.30,\ 
\texttt{sub}:0.20,\ 
\texttt{div}:0.15
\}.
\]

\paragraph{LaTeX expression construction.}
The expression string $e_{\texttt{latex}}$ is deterministically generated using a fixed grammar:
unary operators are rendered as standard LaTeX commands (e.g., $\texttt{log}\mapsto \backslash\texttt{log}(\cdot)$,
$\texttt{sqrt}\mapsto \backslash\texttt{sqrt}\{\cdot\}$, $\texttt{square}\mapsto (\cdot)^2$).
Binary operators are rendered as $+$, $-$, $\backslash\texttt{times}$, or $\backslash\texttt{frac}\{\cdot\}\{\cdot\}$.

\begin{table}[h]
\centering
\small
\setlength{\tabcolsep}{4pt}
\vspace{-5pt}
\caption{An example sample from the \textbf{MathExpr} dataset.}
\begin{tabularx}{\linewidth}{l X}
\toprule
\textbf{Column} & \textbf{Example Value} \\
\midrule
$x_1$ & $2.70$ \\
$x_2$ & $6.40$ \\
operation\_x1 & \texttt{sin} \\
operation\_x2 & \texttt{log} \\
operation\_between & \texttt{mul} \\
latex\_expression &
\texttt{\textbackslash sin(2.70) \textbackslash times \textbackslash log(6.40)} \\
\bottomrule
\end{tabularx}
\label{tab:mathexpr_example}
\vspace{-10pt}
\end{table}

\begin{table}[h]
\centering
\small
\caption{An example sample from the \textbf{ProfileBio} dataset.}
\setlength{\tabcolsep}{4pt}
\begin{tabularx}{\linewidth}{l X}
\toprule
\textbf{Column} & \textbf{Example Value} \\
\midrule
age & $38$ \\
salary & $135$ \\
sex & \texttt{female} \\
birth\_state & \texttt{California} \\
college & \texttt{Harvard University} \\
degree & \texttt{master} \\
occupation & \texttt{software developer} \\
biography &
\texttt{This female individual is in the career-building stage. She was born in California and completed higher education at Harvard University, earning a master degree. She works as a software developer. She earns a strong professional income.} \\
\bottomrule
\end{tabularx}
\label{tab:profilebio_example}
\vspace{-10pt}
\end{table}

\subsubsection{ProfileBio}\label{apdx:profbio}
ProfileBio is a synthetic dataset for evaluating the joint generation of mixed-type personal profiles with a biography text rendered from the structured columns.
Each record contains two numerical columns, five categorical columns, and one text column:
\[
\mathbf{r}
=
(\texttt{age}, \texttt{salary},
 \texttt{sex}, \texttt{birth\_state}, \texttt{college}, \texttt{degree}, \texttt{occupation},
 \texttt{biography}).
\]
The \texttt{biography} field is a natural-language paragraph deterministically constructed from the structured attributes.

Following the setup of MathExpr, we generate $10{,}000$ records and split them into a training/real set and a validation set with a ratio of $9{:}1$. During sampling, each model generates a synthetic dataset with the same cardinality as the real training set. We evaluate distribution fidelity and biography consistency using the metrics described in Appendix~\ref{metric:syntab}.

\paragraph{Categorical value sampling.}
We sample \texttt{sex} uniformly from \{\texttt{male}, \texttt{female}\}. 
The attributes \texttt{birth\_state} and \texttt{college} are sampled from fixed categorical priors over $10$ states and $9$ colleges, respectively. 
The \texttt{degree} attribute is sampled conditionally on \texttt{college}: for \texttt{stanford university} and \texttt{harvard university}, we use
\[
p(\texttt{degree})
=
(0.01, 0.29, 0.40, 0.30),
\]
and for all other colleges, we use
\[
p(\texttt{degree})
=
(0.30, 0.50, 0.15, 0.05),
\]
where the entries correspond to
$(\texttt{associate}, \texttt{bachelor}, \texttt{master}, \texttt{doctoral})$.
Finally, \texttt{occupation} is sampled from a degree-dependent categorical distribution. 
All occupation weights are initialized to $1$. 
For \texttt{doctoral} degrees, the weights of \texttt{research specialist} and \texttt{education professional} are set to $6$ and $4$, respectively; for \texttt{associate} degrees, the weights of \texttt{customer services professional} and \texttt{construction professional} are set to $5$ and $5$, respectively. 
The weights are then normalized into a categorical distribution.

\paragraph{Numerical value sampling.}
We sample \texttt{age} uniformly from integers in $[21,70]$, and draw 
\texttt{salary} from a Gaussian whose mean depends on \texttt{degree}, 
\texttt{occupation}, and \texttt{age}. The degree-dependent base mean is
\[
\mu_0(\texttt{degree})=
\begin{cases}
82,  & \texttt{associate},\\
125, & \texttt{bachelor or master},\\
178, & \texttt{doctoral},
\end{cases}
\]
and, letting $\mathcal{O}^\star=\{\texttt{software developer},\,
\texttt{healthcare practitioner}\}$, the full mean is
\[
\mu \;=\; \mu_0(\texttt{degree})
        + 4\cdot\mathds{1}\{\texttt{occupation}\in\mathcal{O}^\star\} + 0.3\,(\texttt{age}-45).
\]
We then sample $\texttt{salary}\sim\mathcal{N}(\mu,\,5^2)$, round it to 
the nearest integer, and clip it to $[75,200]$.

\paragraph{Biography construction.}
The \texttt{biography} field is deterministically constructed from the structured attributes using a fixed template. 
The \texttt{age} and \texttt{salary} values are first mapped to coarse-grained textual descriptors, and Table~\ref{tab:profbio_template} summarizes both the descriptor mappings and the biography template.

\paragraph{Additional clarification.}
ProfileBio is a fully synthetic dataset constructed from predefined sampling rules and deterministic templates. It does not rely on, copy, or imitate any real individuals. Sensitive personal attributes such as race, religion, health status, or immigration background are intentionally excluded. While attributes such as \texttt{sex}, \texttt{birth\_state}, and \texttt{occupation} are included to evaluate cross-field consistency, the salary generation process is explicitly designed to depend only on \texttt{age}, \texttt{degree}, and \texttt{occupation}, and does not condition on \texttt{sex} or \texttt{birth\_state}. ProfileBio is intended solely as a controlled benchmark for evaluating joint generation fidelity and attribute--text consistency, rather than as a model of real-world socioeconomic distributions.

\begin{table}[!t]
\centering
\small
\vspace{-5pt}
\caption{\textbf{ProfileBio:} Age/salary mapping rules and the biography template.}
\label{tab:profbio_template}
\setlength{\tabcolsep}{6pt}
\renewcommand{\arraystretch}{1.15}
\begin{tabular}{@{}l >{\raggedright\arraybackslash}p{0.74\linewidth}@{}}
\toprule
\textbf{Component} & \textbf{Rule / Template} \\
\midrule
Age descriptor &
[21, 30]: \texttt{in the early career stage} \newline
[31, 40]: \texttt{in the career-building stage} \newline
[41, 50]: \texttt{in the established career stage} \newline
[51, 60]: \texttt{in the advanced career stage} \newline
[61, 70]: \texttt{in the late career stage} \\
\midrule
Salary descriptor &
[75, 100]: \texttt{a comfortable income} \newline
[101, 150]: \texttt{a strong professional income} \newline
[151, 200]: \texttt{a high-level income} \\
\midrule
Biography template &
\ttfamily
This \{sex\} individual is \{age\_desc\}. \{He/She\} was born in \{birth\_state\} and completed higher education at \{college\}, earning a \{degree\} degree. \{He/She\} works as a \{occupation\}. \{He/She\} earns \{salary\_desc\}. \\
\bottomrule
\end{tabular}
\vspace{-10pt}
\end{table}

\subsubsection{Amazon}\label{apdx:amazon}
Amazon is a real-world free-text tabular dataset constructed from the RelBench~\cite{robinson2024relbench} \texttt{rel-amazon} relational benchmark, which contains linked product metadata and user reviews.
We join the \texttt{review} and \texttt{product} tables on \texttt{product\_id} and form a single mixed-type table with two numerical columns, two categorical columns, and six free-form text columns:
\[
\begin{aligned}
(&\texttt{price},\ \texttt{review\_time},\ \texttt{rating},\ \texttt{verified},\ \texttt{category},\ \\
&\texttt{brand},\
\texttt{title},\ \texttt{description},\ \texttt{review\_text},\ \texttt{summary}).
\end{aligned}
\]

We sample $5{,}000$ examples in total and split them into a training/real set and a validation set with a ratio of 9:1.
In addition, we sample another test set of $2{,}250$ examples for downstream utility evaluation.
During sampling, each model generates $4{,}500$ synthetic examples, matching the cardinality of the real training set.
We report distribution fidelity and downstream utility metrics as defined in Appendix~\ref{metric:syntab}.
Table~\ref{tab:amazon_example} shows an example record.

\begin{table}[t]
\centering
\small
\setlength{\tabcolsep}{8pt}
\caption{An example sample from the \textbf{Amazon} dataset.}
\label{tab:amazon_example}
\begin{tabularx}{\linewidth}{l X}
\toprule
\textbf{Column} & \textbf{Example Value} \\
\midrule
price & $5.98$ \\
review\_time & $1970$ \texttt{(days since earliest review date)} \\
rating & $5.0$ \\
verified & \texttt{true} \\
category & \texttt{Science Fiction \& Fantasy > Fantasy} \\
brand & \texttt{Visit Amazon's J. R. R. Tolkien Page} \\
title & \texttt{Hobbit} \\
description & \texttt{The enchanting prelude to "The Lord of the Rings"} \\
review\_text & \texttt{My 13 yr. old grandson was very happy with this book. He likes to know the background of things and this helped him to understand this story.} \\
summary & \texttt{Grandson happy} \\
\bottomrule
\end{tabularx}
\vspace{-10pt}
\end{table}

\subsubsection{Arxiv}\label{apdx:arxiv}

Arxiv is another real-world free-text tabular dataset constructed from the RelBench \texttt{rel-arxiv} relational benchmark, which contains metadata and abstracts of arXiv papers.
We construct a single mixed-type table at the paper level with four numerical columns, one categorical column, and three free-form text columns:
\[
\begin{aligned}
(&\texttt{submission\_time},\ \texttt{submission\_year},\ \texttt{title\_length}, \\
&\texttt{abstract\_length},\
 \texttt{category},\ \texttt{title},\ \texttt{arxiv\_code},\ \texttt{abstract}).
\end{aligned}
\]

We sample $4{,}500$ examples for training and validation and split them into a real training set and a validation set with a ratio of 8:1.
In addition, we sample another test set of $2{,}000$ examples for downstream utility evaluation. 
During sampling, each model generates a synthetic dataset with the same cardinality as the real training set. 
We report distribution fidelity and downstream utility metrics as defined in Appendix~\ref{metric:syntab}. 
Table~\ref{tab:arxiv_example} shows an example record.

\begin{table}[t]
\vspace{-5pt}
\centering
\small
\setlength{\tabcolsep}{8pt}
\caption{An example record from the \textbf{Arxiv} dataset.}
\label{tab:arxiv_example}
\begin{tabularx}{\linewidth}{l X}
\toprule
\textbf{Column} & \textbf{Example Value} \\
\midrule
submission\_time & $1672$ \texttt{(days since earliest submission date in the dataset)} \\
submission\_year & $2022$ \\
title\_length & $7$ \\
abstract\_length & $90$ \\
category & \texttt{hep-ph} \\
title & \texttt{IRC-safe Graph Autoencoder for unsupervised anomaly detection} \\
arxiv\_code & \texttt{arXiv:2204.12231} \\
abstract & \texttt{Anomaly detection through employing machine learning techniques has emerged as a novel powerful tool in the search for new physics beyond the Standard Model. Historically similar to the development of jet observables, theoretical consistency has not always assumed a central role in the fast development of algorithms and neural network architectures. In this work, we construct an infrared and collinear safe autoencoder based on graph neural networks by employing energy-weighted message passing. We demonstrate that whilst this approach has theoretically favourable properties, it also exhibits formidable sensitivity to non-QCD structures.} \\
\bottomrule
\end{tabularx}
\vspace{-10pt}
\end{table}

\subsubsection{Real-world Tabular Datasets}\label{apdx:real-tabular}
We evaluate on five widely-used real-world tabular datasets from the UCI Machine Learning Repository.\footnote{\texttt{https://archive.ics.uci.edu/datasets}}
These datasets cover both classification and regression tasks, providing a diverse testbed for regular tabular generation with only numerical and categorical features.
Specifically, Adult, Default, Shoppers, and Magic are used for classification, while Beijing is used for regression.
Dataset statistics and the corresponding train/validation/test splits are summarized in Table~\ref{tab:uci_stats}.

\begin{table}[t]
\centering
\small
\setlength{\tabcolsep}{6pt}
\caption{Statistics of real-world tabular datasets. \#Num denotes the number of numerical columns, and \#Cat denotes the number of categorical columns. \#Max Cat is the maximum number of categories among all categorical columns.}
\label{tab:uci_stats}
\begin{tabular}{lcccccccl}
\toprule
\textbf{Dataset} & \textbf{\#Rows} & \textbf{\#Num} & \textbf{\#Cat} & \textbf{\#Max Cat} & \textbf{\#Train} & \textbf{\#Validation} & \textbf{\#Test} & \textbf{Task} \\
\midrule
\textbf{Adult} & $48{,}842$ & $6$ & $9$ & $42$ & $28{,}943$ & $3{,}618$ & $16{,}281$ & Classification \\
\textbf{Default} & $30{,}000$ & $14$ & $11$ & $11$ & $24{,}000$ & $3{,}000$ & $3{,}000$ & Classification \\
\textbf{Shoppers} & $12{,}330$ & $10$ & $8$ & $20$ & $9{,}864$ & $1{,}233$ & $1{,}233$ & Classification \\
\textbf{Magic} & $19{,}019$ & $10$ & $1$ & $2$ & $15{,}215$ & $1{,}902$ & $1{,}902$ & Classification \\
\textbf{Beijing} & $43{,}824$ & $7$ & $5$ & $31$ & $35{,}058$ & $4{,}383$ & $4{,}383$ & Regression \\
\bottomrule
\end{tabular}
\vspace{-10pt}
\end{table}

\subsection{Metrics}\label{apdx:metrics}
\subsubsection{Shape and Trend}
We evaluate distribution fidelity on all datasets using two general-purpose metrics: \textbf{Shape} and \textbf{Trend}.
Shape and Trend are adopted from SDMetrics\footnote{\url{https://docs.sdv.dev/sdmetrics}}, which quantify the marginal column-wise similarity and the pairwise dependency preservation between real and synthetic data, respectively.

\textbf{Shape.} \textit{Kolmogorov-Smirnov Test (KST):} 
This metric quantifies the alignment between the real distribution $p_r(x)$ and the synthetic distribution $p_s(x)$ by calculating the maximum divergence between their respective Cumulative Distribution Functions (CDFs):
\begin{equation}
    \text{KST} = \sup_{x} |F_r(x) - F_s(x)|,
\end{equation}
where $F_r(x)$ and $F_s(x)$ denote the CDFs derived from the probability densities:
\begin{equation}
    F(x) = \int_{-\infty}^{x} p(t)\mathrm{d}t.
\end{equation}

\textit{Total Variation Distance (TVD):} 
For categorical variables, we evaluate the discrepancy in probability mass using the TVD. It is defined as half the sum of the absolute differences between the category frequencies observed in the real data, $R(\omega)$, and the synthetic data, $S(\omega)$:
\begin{equation}
    \text{TVD} = \frac{1}{2} \sum_{\omega \in \Omega} |R(\omega) - S(\omega)|,
\end{equation}
where $\Omega$ represents the set of all possible categories within a given column.

\textbf{Trend.} \textit{Pearson Correlation Score:} 
We examine the preservation of linear dependencies between continuous columns using the Pearson correlation coefficient, $\rho_{x,y}$, defined as:
\begin{equation}
    \rho_{x,y} = \frac{\text{Cov}(x,y)}{\sigma_x \sigma_y},
\end{equation}
where $\text{Cov}$ represents covariance and $\sigma$ denotes the standard deviation. To evaluate the overall preservation of these trends, we compute the Pearson Score as the normalized average absolute error between the correlation matrices of the real ($\rho^R$) and synthetic ($\rho^S$) datasets:
\begin{equation}
    \text{Pearson Score} = \frac{1}{2} \mathbb{E}_{x,y} |\rho^R(x,y) - \rho^S(x,y)|.
\end{equation}
Since $\rho \in [-1, 1]$, the factor of $1/2$ normalizes the score to the range $[0, 1]$, where a lower score indicates superior correlation preservation.

\textit{Contingency Similarity:} 
To measure the consistency of pairwise associations between categorical columns $A$ and $B$, we utilize a metric based on the Total Variation Distance applied to contingency tables. The Contingency Score is calculated as:
\begin{equation}
    \text{Contingency Score} = \frac{1}{2} \sum_{\alpha \in A} \sum_{\beta \in B} |R_{\alpha,\beta} - S_{\alpha,\beta}|,
\end{equation}
where $R_{\alpha,\beta}$ and $S_{\alpha,\beta}$ correspond to the joint frequencies of category pair $(\alpha, \beta)$ in the real and synthetic datasets, respectively.

\subsubsection{$\alpha$-Precision}
Following~\citet{alaa2022faithful}, $\alpha$-Precision is a sample-level fidelity 
metric that measures whether each synthetic sample falls within the $\alpha$-support of the real data distribution, i.e., the smallest region containing an $\alpha$ fraction of the real probability mass. Intuitively, it quantifies how typical synthetic samples are with respect to the real distribution rather than lying in low-density or out-of-distribution regions. A higher $\alpha$-Precision indicates higher fidelity of the generated samples.

\subsubsection{Detection Score}
The Detection Score is computed via the Classifier Two-Sample Test (C2ST) implemented in SDMetrics, where a logistic regression classifier is trained to distinguish real samples from synthetic ones. The score is derived from the classifier's misclassification rate: when the synthetic distribution closely matches the real one, the classifier performs near chance level, yielding a higher Detection Score. We report the Detection Score in $[0,1]$, with higher values indicating that synthetic samples are statistically indistinguishable from real samples.

\subsubsection{Machine Learning Efficiency}
For datasets with an associated downstream prediction task, we additionally report Machine Learning Efficiency (MLE) to assess utility by comparing the test performance of models trained on real versus synthetic samples.
In particular, we apply MLE to the regular real-world tabular datasets.
During evaluation, we train an XGBoost model on the real training set (further split with an 8:1 ratio for validation and hyperparameter tuning) and evaluate it on a held-out real test set.
We then train an identical model on the synthetic dataset and evaluate it on the same real test set.
We report the test performance of the model trained on synthetic data (AUC for classification and RMSE for regression); the closer it is to that of the model trained on real data, the better synthetic data can serve as a substitute for real data in downstream predictive modeling. The Average Gap in Table~\ref{tab:mle} is the relative drop in test performance when training on synthetic instead of real data, averaged over the datasets.

\subsubsection{Distance Closest Record (DCR)}
DCR evaluates whether the generative model memorizes training samples 
rather than learning the underlying distribution, serving as a data privacy metric. For each synthetic sample, we compute its distance to the 
nearest record in both the training set and a held-out test set, and report 
the proportion of synthetic samples whose nearest neighbor lies in the 
training set. A score close to $50\%$ is ideal: it indicates that synthetic 
samples are equally close to training and test data, suggesting the model 
has captured the distribution rather than reproducing training instances.

\subsubsection{Dataset-specific Metrics}\label{metric:syntab}
\textbf{MathExpr.}
Beyond standard distribution metrics (Shape and Trend), we evaluate whether the generated LaTeX expression is structurally and numerically consistent with the structured number and operation columns.
We report:
(1) Operation Match Rate (Op-MR), which verifies if the unary/binary operator tokens implied by the expression $e_{\texttt{latex}}$ strictly align with the structured operations $(o_1,o_2,o_3)$;
and (2) Expression match rate (Exp-MR), which evaluates the joint validity of structure and values.
Specifically, a generated expression is considered a match only if it (i) satisfies operation correctness (as in Op-MR) and (ii) contains two numeric literals whose values align with the structured fields $(x_1, x_2)$ up to a small relative tolerance. Concretely, letting $\hat{x}_1,\hat{x}_2$ be the literals extracted from the generated LaTeX string, we require $\lvert \hat{x}_i - x_i \rvert / x_i \le \delta$ for $i\in\{1,2\}$. We fix $\delta = 0.07$ to make the metric robust to minor numeric drift that can arise from stochastic generation and continuous-value approximation (e.g., $0.29$ vs.\ $0.30$), while still penalizing outputs whose numeric content is meaningfully inconsistent with the structured inputs.

\textbf{ProfileBio.}
For ProfileBio, we assess cross-modality consistency between structured attributes and the generated biography text.
We compute a rule-based Match Rate by checking whether the biography instantiates the required template slots with values consistent with the corresponding structured fields (see the example template in Table~\ref{tab:profbio_template}).
Concretely, for each record we verify that the text reflects core attributes (e.g., \texttt{sex}, \texttt{birth\_state}, \texttt{college}, \texttt{degree}, \texttt{occupation}) as well as the derived descriptors for continuous variables (\texttt{Age} and \texttt{Salary}).
Since mapping continuous values into discrete natural-language descriptors introduces semantic fuzziness near bin boundaries, we apply a small boundary relaxation tolerance $\delta=0.06$ when validating the age/salary descriptors.
This avoids rigid thresholding artifacts where values close to a boundary may legitimately share descriptions from neighboring categories; for instance, age $30.5$ can reasonably be described as either ``in the early career stage'' or ``in the career-building stage.''

\textbf{Amazon.}
For the Amazon dataset, we measure downstream utility via an MLE-like protocol, with \texttt{rating} as the target label in $\{1,2,3,4,5\}$ (treated as a multi-class classification task).
We first serialize all non-numerical fields into structured text and encode them using the sentence embedding model Nomic.
We then train an XGBoost classifier using either (i) text embeddings only or (ii) text embeddings concatenated with numerical features, which helps isolate the contribution of generated text/categorical fields.
We report Macro-AUC on the held-out test set as the utility metric.

\textbf{Arxiv.}
Similar to the Amazon dataset, we measure downstream utility via an MLE-like protocol, with \texttt{Category} as the target label (treated as a multi-class classification task). The original \texttt{rel-arxiv} benchmark contains more than $50$ paper categories, which would induce severe class imbalance and weaken the signal of cross-field consistency; we therefore restrict our benchmark to samples from the $15$ most frequent categories, yielding a relatively balanced $15$-way classification task. We serialize all non-numerical fields into structured text and encode them using the sentence embedding model Nomic. We then train an XGBoost classifier using either (i) text embeddings only or (ii) text embeddings concatenated with numerical features. For each configuration, Macro-AUC on the held-out test is reported as the utility metric.

\section{Implementation Details}\label{apdx:impl}
We implement \name based on PyTorch~\cite{paszke2019pytorch}; the code is available at \url{https://github.com/ilikevegetable/TabDLM}. All experiments are run on a single NVIDIA A100-80GB GPU.

\textbf{Data preprocessing.}
We follow the same preprocessing method as the prior tabular diffusion model TabDiff~\cite{tabdiff}.
Missing numerical values are imputed with the column mean, and missing categorical values are treated as an additional category.
To stabilize optimization across heterogeneous numerical scales, we apply a quantile-based transformation to numerical columns during training and invert the transform after sampling to recover values in the original space.

\textbf{Data splits.}
We adopt the same data split protocol as the TabDiff setting: each dataset is partitioned into a real set and a test set.
Models are trained on the real set.
For downstream utility evaluation, we further split the real set into training and validation subsets and reserve the test set strictly for evaluation.

\textbf{Architecture.}
We build \name{} on top of the LLaDA-8B base model for all experiments.
To incorporate numerical channels, each scalar value is first mapped into a $d$-dimensional latent using a lightweight pretrained float encoder, implemented as a 3-layer MLP with hidden width $\lfloor \sqrt{d}\rfloor$ and SiLU activations~\cite{elfwing2018sigmoid}.
A float decoder (LayerNorm + linear projection) maps the latent back to a scalar.
We set the numerical latent dimension to $d=512$ by default and keep the pretrained float encoder/decoder frozen during joint training.
To interface numerical latents with the MDLM embedding space, we further apply a two-stage projection: an \emph{input projector} that maps per-feature latents from $d$ to the MDLM hidden size $D$ using a 2-layer MLP ($d\!\rightarrow\!1024\!\rightarrow\!D$) with SiLU and dropout (with LayerNorm), and an \emph{output projector} that maps MDLM hidden states back to the numerical latent space via a symmetric 2-layer MLP ($D\!\rightarrow\!1024\!\rightarrow\!d$). The dimension $D$ in LLaDA-8B is 4096.

\textbf{Hyperparameter settings.}
We fine-tune models with LoRA.
Unless otherwise stated, we use LoRA rank $r=16$, scaling factor $\alpha=32$, and dropout $0.05$, and apply LoRA to the attention and MLP components of Transformer blocks.  All models are trained and evaluated on a single NVIDIA A100-80GB GPU; training time, memory, and sampling cost are reported in Tables~\ref{tab:cost} and~\ref{tab:wallclock}.
Across methods, we keep the optimizer and training recipe identical; the only dataset-dependent choice is the number of training epochs due to varying dataset sizes.
Specifically, we train for 10 epochs on Adult and Beijing, 30 epochs on Shoppers, 15 epochs on Magic and Default, and 75 epochs on MathExpr, Amazon, Arxiv, and ProfileBio. We use AdamW with learning rate $2\times 10^{-4}$, warmup ratio $0.1$, $(\beta_1,\beta_2)=(0.9,0.98)$, weight decay $10^{-4}$, and $\epsilon=10^{-8}$.
All experiments use the same fixed random seed and bf16 training when enabled.

\textbf{Noise schedule.}
For the continuous numerical diffusion, we adopt the per-feature 
power-mean noise schedule following TabDiff~\citep{tabdiff}, which 
extends the EDM parameterization~\citep{karras2022elucidating} with 
a learnable shape parameter $\rho_i$ for each numerical column. 
The schedule smoothly interpolates between $\sigma_{\min}$ and 
$\sigma_{\max}$ over normalized time $t\in[0,1]$; we set 
$\sigma_{\min}=0.002$ and $\sigma_{\max}=80.0$ following 
EDM defaults.
Formally, for each numerical feature $i\in\{1,\ldots,M_{\text{num}}\}$:
\begin{equation}
\sigma^{\text{num}}_{\rho_i}(t) =
\left(
\sigma_{\min}^{1/\rho_i} + t\left(
\sigma_{\max}^{1/\rho_i}-\sigma_{\min}^{1/\rho_i}
\right)
\right)^{\rho_i}.
\end{equation}
Learning $\rho_i$ per column accommodates heterogeneous numerical 
distributions and yields a better-conditioned diffusion process 
than a single global $\rho$.

\begin{algorithm}[t]
\caption{Sampling}
\label{alg:tabdlm_sampling}
\begin{algorithmic}[1]
\Require Schema prompt $\mathbf{p}$; generation length $G$;
         number of reverse steps $T$;
         noise schedules $\{\sigma_t\}_{t=0}^{T}$ and 
         churned schedule $\{\hat{\sigma}_t\}_{t=0}^{T}$;
         remasking policy $\pi\!\in\!\{\textsc{HighConf},\textsc{Rand}\}$.
\Ensure  Synthetic record  
    $\mathbf{x}=\big(x_j\big)_{j\in\mathcal{I}_{\text{num}}\cup\mathcal{I}_{\text{cat}}\cup\mathcal{I}_{\text{text}}}$.
\vspace{2pt}
\State \textbf{// Initialization}
\State $\mathbf{z}_{\mathbf{p}} \gets \text{EMB}(\mathbf{p})$
\State $\mathbf{x}_{T}^{\text{tok}} \gets \texttt{[MASK]}^{G}$
       \Comment{discrete state for $\mathcal{I}_{\text{cat}}\!\cup\!\mathcal{I}_{\text{text}}$}
\State $\mathbf{x}_{T}^{\text{num}} \sim \mathcal{N}\!\big(\mathbf{0},\,\sigma_{\max}^{2}\mathbf{I}\big)$
       \Comment{initial state from a Gaussian prior}
\vspace{2pt}
\For{$t = T, T-1, \dots, 1$}
    \State \textbf{// (i) Numerical perturbation (EDM churn)}
    \State $\boldsymbol{\epsilon}\sim\mathcal{N}(\mathbf{0},\mathbf{I})$;\quad
           $\hat{\mathbf{x}}_{t}^{\text{num}} \gets \mathbf{x}_{t}^{\text{num}} 
           + \sqrt{\hat{\sigma}_{t}^{2}-\sigma_{t}^{2}}\;\boldsymbol{\epsilon}$
    \vspace{1pt}
    \State \textbf{// (ii) Map both modalities to MDLM token embeddings}
    \State $\mathbf{z}_{t}^{\text{num}} \gets 
           \text{PROJ}_{e}\!\big(\text{ENC}(\hat{\mathbf{x}}_{t}^{\text{num}})\big)$
           \Comment{Eq.~\eqref{eq:8}}
    \State $\mathbf{z}_{t}^{\text{tok}} \gets \text{EMB}(\mathbf{x}_{t}^{\text{tok}})$
           \Comment{Eq.~\eqref{eq:9}}
    \State $\mathbf{z}_{t} \gets 
           \big[\mathbf{z}_{\mathbf{p}};\, \mathbf{z}_{t}^{\text{tok}};\, \mathbf{z}_{t}^{\text{num}}\big]$
    \vspace{1pt}
    \State \textbf{// (iii) Joint bidirectional denoising}
    \State $\mathbf{o}_{t} = \big[\mathbf{o}_{t}^{\text{p}};\, \mathbf{o}_{t}^{\text{tok}};\, \mathbf{o}_{t}^{\text{num}}\big] \gets              \text{MDLM}(\mathbf{z}_{t},\,t)$
           \Comment{Eq.~\eqref{eq:10}}
    \vspace{1pt}
    \State \textbf{// (iv) Discrete reverse step: progressive unmasking}
    \State $p_{\theta}(\cdot\mid\mathbf{x}_{t}^{\text{tok}}) \gets 
           \text{HEAD}(\mathbf{o}_{t}^{\text{tok}})$
           \Comment{Eq.~\eqref{eq:12}}
    \State $\tilde{\mathbf{x}}_{0}^{\text{tok}} \sim p_{\theta}(\cdot\mid\mathbf{x}_{t}^{\text{tok}})$
           \Comment{sample candidate tokens}
    \State $\mathbf{x}_{t-1}^{\text{tok}} \gets 
           \textsc{Unmask}_{\pi}\!\big(\mathbf{x}_{t}^{\text{tok}},\,
           \tilde{\mathbf{x}}_{0}^{\text{tok}},\,p_{\theta},\,t\big)$
           \Comment{reveal $G/T$ positions per step}
    \vspace{1pt}
    \State \textbf{// (v) Continuous reverse step: EDM--Euler update}
    \State $\tilde{\mathbf{x}}_{t}^{\text{num}} \gets 
           \text{DEC}\!\big(\text{PROJ}_{d}(\mathbf{o}_{t}^{\text{num}})\big)$
           \Comment{Eq.~\eqref{eq:11}}
    \State $\mathbf{d}_{t} \gets 
           \big(\hat{\mathbf{x}}_{t}^{\text{num}}-\tilde{\mathbf{x}}_{t}^{\text{num}}\big)
           \big/\hat{\sigma}_{t}$
    \State $\mathbf{x}_{t-1}^{\text{num}} \gets 
           \hat{\mathbf{x}}_{t}^{\text{num}} + 
           \big(\sigma_{t-1}-\hat{\sigma}_{t}\big)\,\mathbf{d}_{t}$
           \Comment{discretization of Eq.~\eqref{eq:numerical_backward}}
\EndFor
\vspace{2pt}
\State \textbf{// Finalization}
\State $\big\{x_{j}\big\}_{j\in\mathcal{I}_{\text{cat}}\cup\mathcal{I}_{\text{text}}} 
       \gets \text{Detokenize}(\mathbf{x}_{0}^{\text{tok}})$
\State $\big\{x_{j}\big\}_{j\in\mathcal{I}_{\text{num}}} 
       \gets \text{Denorm}(\mathbf{x}_{0}^{\text{num}})$
       \Comment{numerical denormalization}
\State \Return $\mathbf{x}$
\end{algorithmic}
\end{algorithm}

\textbf{Sampling.}\label{apdx:sample_detail}
\name{} performs joint sampling over discrete tokens 
(categorical/text) and continuous numerical features through a coupled 
reverse process discretized into $T$ steps; the full procedure is 
summarized in Algorithm~\ref{alg:tabdlm_sampling}. At each step, we 
(i) inject EDM-style churn noise~\citep{karras2022elucidating} into the 
numerical state, (ii) embed both modalities into a shared MDLM input by 
mapping numerical values through the number encoder and tokens through 
the embedding table, (iii) run a single bidirectional MDLM forward pass 
that jointly conditions text denoising on noisy numericals and vice 
versa, and (iv) apply modality-specific reverse updates: progressive 
unmasking on the discrete side and an EDM--Euler step on the continuous 
side.

For discrete unmasking we consider two policies: 
(\textit{i})~\textbf{high-confidence} unmasking, which reveals positions 
with the highest predicted token probability, and 
(\textit{ii})~\textbf{random} unmasking, which reveals a uniformly 
random subset. In both cases we reveal $G/T$ tokens per step, ensuring 
a smooth transition from a fully-masked sequence to a fully-specified 
one. After the final step, we decode tokens through the LM head and denormalize the numerical values to obtain the final mixed-type sample.

\textbf{\names.}\label{apdx:cost}
To test whether our formulation carries over to a different backbone and to measure the trade-off between text quality and efficiency, we instantiate the same framework on the 170M-parameter masked diffusion language model of~\citet{sahoo2024simple}, which is pretrained on OpenWebText with the GPT-2 tokenizer. The number encoder and decoder, the two projectors and the training objective are identical to those of \name; besides the backbone, \names differs in that, given its size, all parameters are fine-tuned instead of using LoRA (AdamW, learning rate $4\times10^{-4}$, batch size 64).

\textbf{Computational cost.}
Table~\ref{tab:cost} reports the training and sampling cost on Shoppers, and Table~\ref{tab:wallclock} the sampling time per generated row on Amazon, all measured on a single NVIDIA A100-80GB GPU. Each of the $T$ reverse steps (Algorithm~\ref{alg:tabdlm_sampling}) is a single forward pass of the backbone over the full sequence, i.e., the prompt, the $G$ token positions, and one position per numerical value, and updates the text and the numerical values together. Generating a row therefore takes $T$ forward passes, where $T$ is a hyperparameter set independently of $G$, since $G/T$ tokens are unmasked per step. Because each numerical value occupies a single position rather than a sequence of digit tokens, the sequence stays short, which keeps each forward pass cheap on wide tables. \name is fine-tuned with LoRA ($r{=}16$, $\alpha{=}32$ on the attention and MLP blocks), i.e., 41.9M trainable parameters on top of the frozen 8B backbone.
Specialized tabular diffusion models such as TabDiff are the cheapest to sample per generated row, but they cannot generate free-form text. \names trains faster than TabDiff with less memory and samples faster than both Qwen2.5-7B baselines on Amazon, while staying close to \name on average on the tables of Table~\ref{tab:shape_trend}; our formulation therefore largely does not depend on a large backbone for structured fidelity. \name still improves on \names in Shape and Trend on Amazon and Arxiv and in Bio-MR on ProfileBio (Tables~\ref{tab:mathexpr_profilebio} and~\ref{tab:amazon_arxiv}), which is why we adopt LLaDA-8B by default. Finally, \name and LLaDA-8B$_{\text{SFT}}$ cost the same at the same number of steps, so the sampling time of \name is that of its backbone, and acceleration techniques for diffusion language models such as KV caching~\cite{wu2025fast} are compatible with our formulation.

\begin{table}[h]
\centering
\footnotesize
\setlength{\tabcolsep}{5pt}
\caption{Training and sampling cost on Shoppers. Memory is measured under each method's default training batch size (bs); the sampling time of \names and \name is per step at batch size 16.}
\label{tab:cost}
\begin{tabular}{lcccc}
\toprule
\textbf{Method} & \textbf{Backbone / trainable} & \textbf{Train time} & \textbf{Train memory} & \textbf{Sampling} \\
\midrule
TabDiff & $10.6$M\,/\,$10.6$M & $1.0$\,h & $72$\,GB (bs $1000$) & $0.072$\,s/step\,$\times\,50$ \\
\names  & $170$M\,/\,$170$M   & $0.77$\,h & $20$\,GB (bs $64$) & $0.017$\,s/step\,$\times\,48$ \\
\name   & $8$B\,/\,$41.9$M (LoRA) & $9.0$\,h & $54$\,GB (bs $32$) & $0.39$\,s/step\,$\times\,48$ \\
\bottomrule
\end{tabular}
\end{table}

\begin{table}[h]
\centering
\footnotesize
\setlength{\tabcolsep}{6pt}
\caption{Sampling time per generated row on Amazon. Specialized tabular diffusion models are absent because they cannot generate free-form text.}
\vspace{5pt}
\label{tab:wallclock}
\begin{tabular}{lcc}
\toprule
\textbf{Method} & \textbf{Sampling steps} & \textbf{Time / row} \\
\midrule
GReaT (124M) & -- & $\sim0.05$\,s \\
Qwen2.5-7B$_{\text{ICL}}$ & -- & $\sim0.22$\,s \\
Qwen2.5-14B$_{\text{ICL}}$ & -- & $\sim0.34$\,s \\
Qwen2.5-7B$_{\text{SFT}}$ & -- & $\sim0.38$\,s \\
LLaDA-8B$_{\text{SFT}}$ & $128$ & $\sim3.8$\,s \\
\names & $128$ & $\sim0.17$\,s \\
\name & $128$ & $\sim3.8$\,s \\
\bottomrule
\end{tabular}
\end{table}

\section{Additional Experimental Results}\label{apdx:add_exps}

\subsection{Full results of Shape and Trend Metrics}
\label{apdx:full_shape_trend}
Shape and Trend are fidelity metrics measuring column-wise marginal distributions and pair-wise column dependencies, respectively. Tables~\ref{tab:shape} and~\ref{tab:trend} report the complete results on real-world tabular datasets without free-form text features.

\begin{table*}[htbp]
\centering
\footnotesize
\setlength{\tabcolsep}{6pt}
\caption{Per-dataset expansion of the Shape entries of Table~\ref{tab:shape_trend}. Performance comparison on the error rates (\%) of \textbf{Shape} $\downarrow$. (\redbf{B}: best overall; \textbf{B}: best in language-based model) }
\vspace{-5pt}
\label{tab:shape}
\begin{tabular}{lccccc|>{\columncolor{gray!15}}c}
\toprule
\textbf{Method} & \textbf{Adult} & \textbf{Default} & \textbf{Shoppers} & \textbf{Magic} & \textbf{Beijing} & \textbf{Average} \\
\midrule
CTGAN   & $16.84\pm0.03$ & $16.83\pm0.04$ & $21.15\pm0.10$ & $9.81\pm0.08$  & $21.39\pm0.05$ & $17.20$ \\
TVAE    & $14.22\pm0.08$ & $10.17\pm0.05$ & $24.51\pm0.06$ & $8.25\pm0.06$  & $19.16\pm0.06$ & $15.26$ \\
GOGGLE  & $16.97$        & $17.02$        & $22.33$        & $1.90$         & $16.93$        & $15.03$ \\
STaSy   & $11.29\pm0.06$ & $5.77\pm0.06$  & $9.37\pm0.09$  & $6.29\pm0.13$  & $6.71\pm0.03$  & $7.89$ \\
CoDi    & $21.38\pm0.06$ & $15.77\pm0.07$ & $31.84\pm0.05$ & $11.56\pm0.26$ & $16.94\pm0.02$ & $19.50$ \\
TabDDPM & $1.75\pm0.03$  & $1.57\pm0.08$  & $2.72\pm0.13$  & $1.01\pm0.09$  & $1.30\pm0.03$  & $1.67$ \\
TabSyn  & $0.81\pm0.05$  & $1.01\pm0.08$ & $1.44\pm0.07$ & $1.03\pm0.14$ & $1.26\pm0.05$ & $1.11$ \\
TabDiff & \redbf{$\mathbf{0.63\pm0.05}$} & $1.24\pm0.07$  & $1.28\pm0.09$  & $0.78\pm0.08$  & $1.03\pm0.05$ & $0.99$ \\
TabNAT & $0.73$ & \redbf{$\mathbf{0.80}$}  & \redbf{$\mathbf{1.11}$}  & \redbf{$\mathbf{0.62}$}  & \redbf{$\mathbf{0.79}$} & \redbf{$\mathbf{0.81}$} \\
\midrule
GReaT   & $12.12\pm0.04$ & $19.94\pm0.06$ & $14.51\pm0.12$ & $16.16\pm0.09$ & $8.25\pm0.12$  & $14.20$ \\
DiffLM & $9.74$ & $9.06$ & $10.07$ & $7.53$ & $6.35$ & $8.55$ \\
Qwen2.5-7B$_{\text{ICL}}$ & $27.80$ & $22.39$ & $34.37$ & $13.41$ & $31.83$ & $25.96$ \\
Qwen2.5-14B$_{\text{ICL}}$
& $25.17$ & $21.52$ & $51.21$ & $17.51$ & $29.76$ & $29.03$ \\
Qwen2.5-7B$_{\text{SFT}}$ & $5.11$ & $2.58$ & $8.76$ & $7.53$ & $7.49$ & $6.29$ \\
LLaDA-8B$_{\text{SFT}}$ & $1.69$ & $2.00$ & $13.56$ & $5.60$ & $3.64$ & $5.30$ \\
\midrule
\textbf{\names} & $1.49$ & $3.64$ & $4.54$ & $\mathbf{1.96}$ & $\mathbf{2.39}$ & $2.80$ \\
\textbf{\name} & $\mathbf{1.46}$ & $\mathbf{1.18}$ & $\mathbf{2.05}$ & $2.93$ & $2.42$ & $\mathbf{2.01}$ \\
\bottomrule
\end{tabular}
\end{table*}

\begin{table*}[htbp]
\centering
\footnotesize
\caption{Per-dataset expansion of the Trend entries of Table~\ref{tab:shape_trend}. Performance comparison on the error rates (\%) of \textbf{Trend} $\downarrow$.(\redbf{B}: best overall; \textbf{B}: best in language-based model)}
\vspace{-5pt}
\setlength{\tabcolsep}{6pt}
\label{tab:trend}
\begin{tabular}{lccccc|>{\columncolor{gray!15}}c}
\toprule
\textbf{Method} & \textbf{Adult} & \textbf{Default} & \textbf{Shoppers} & \textbf{Magic} & \textbf{Beijing} & \textbf{Average} \\
\midrule
CTGAN   & $20.23\pm1.20$ & $26.95\pm0.93$ & $13.08\pm0.16$ & $7.00\pm0.19$  & $22.95\pm0.08$ & $18.04$ \\
TVAE    & $14.15\pm0.88$ & $19.50\pm0.95$ & $18.67\pm0.38$ & $5.82\pm0.49$  & $18.01\pm0.08$ & $15.23$ \\
GOGGLE  & $45.29$        & $21.94$        & $23.90$        & $9.47$         & $45.94$        & $29.31$ \\
STaSy   & $14.51\pm0.25$ & $5.96\pm0.26$  & $8.49\pm0.15$  & $6.61\pm0.53$  & $8.00\pm0.10$  & $8.71$ \\
CoDi    & $22.49\pm0.08$ & $68.41\pm0.05$ & $17.78\pm0.11$ & $6.53\pm0.25$  & $7.07\pm0.15$  & $24.46$ \\
TabDDPM & $3.01\pm0.25$  & $4.89\pm0.10$  & $6.61\pm0.16$  & $1.70\pm0.22$  & $2.71\pm0.09$  & $3.78$ \\
TabSyn  & $1.93\pm0.07$  & $2.81\pm0.48$  & $2.13\pm0.10$  & $0.88\pm0.18$  & $3.13\pm0.34$  & $2.18$ \\
TabDiff & \redbf{$\mathbf{1.49\pm0.16}$} & $2.55\pm0.75$ & $1.74\pm0.08$ & \redbf{$\mathbf{0.76\pm0.12}$} & $2.59\pm0.15$ & \redbf{$\mathbf{1.83}$} \\
TabNAT & $1.61$ & \redbf{$\mathbf{2.21}$} & \redbf{$\mathbf{1.67}$} & $1.37$ & \redbf{$\mathbf{2.48}$} & $1.87$ \\
\midrule
GReaT   & $17.59\pm0.22$ & $70.02\pm0.12$ & $45.16\pm0.18$ & $10.23\pm0.40$ & $59.60\pm0.55$ & $40.52$ \\
Qwen2.5-7B$_{\text{ICL}}$ & $37.61$ & $31.02$ & $34.77$ & $20.62$ & $40.83$ & $32.97$\\
Qwen2.5-14B$_{\text{ICL}}$
& $42.67$ & $28.72$ & $42.32$ & $18.11$ & $38.20$ & $34.00$ \\
Qwen2.5-7B$_{\text{SFT}}$ & $17.03$ & $16.99$ & $10.28$ & $14.45$ & $28.68$ & $17.49$ \\
LLaDA-8B$_{\text{SFT}}$ & $26.15$ & $26.30$ & $16.42$ & $13.71$ & $27.90$ & $22.10$ \\
\midrule
\textbf{\names} & $\mathbf{2.56}$ & $6.19$ & $4.62$ & $\mathbf{1.76}$ & $\mathbf{2.71}$ & $3.57$ \\
\textbf{\name} & $2.74$ & $\mathbf{2.33}$ & $\mathbf{2.40}$ & $2.85$ & $2.93$ & $\mathbf{2.65}$ \\
\bottomrule
\end{tabular}
\end{table*}

\subsection{Evaluation results of MLE for real-world tabular dataset}
\label{apdx:real_world_mle}
In this section, we provide the evaluation results of MLE for real-world tabular datasets without free-form text features in Table~\ref{tab:mle}. As shown in the table, \name achieves competitive MLE performance on Adult, Default, Shoppers, and Magic, demonstrating its ability to faithfully capture the underlying data distribution and support high-quality synthetic data for downstream training. Notably, \name attains the best overall score on Shoppers, and on Default even surpasses models trained on real data, suggesting that \name can produce samples that are both distributionally consistent and beneficial for improving generalization. Among language-based models, \names attains the lowest average gap. We observe a noticeable performance drop on Beijing, which may be caused by the large fraction of missing values in the target column, where our mean-imputation preprocessing could distort the true target distribution and hurt likelihood-based metrics such as MLE.

\begin{table*}[htbp]
\centering
\footnotesize
\caption{Evaluation of MLE. 
AUC is used for classification tasks and RMSE for regression tasks. (\redbf{B}: best overall; \textbf{B}: best in language-based model)}
\vspace{-5pt}
\label{tab:mle}
\setlength{\tabcolsep}{6pt}
\begin{tabular}{lccccc>{\columncolor{gray!15}}c}
\toprule
\multirow{2}{*}{\textbf{Methods}}
& \multicolumn{1}{c}{\textbf{Adult}}
& \multicolumn{1}{c}{\textbf{Default}}
& \multicolumn{1}{c}{\textbf{Shoppers}}
& \multicolumn{1}{c}{\textbf{Magic}}
& \multicolumn{1}{c}{\textbf{Beijing}}
& \multicolumn{1}{>{\columncolor{gray!15}}c}{\textbf{Average Gap}} \\
\cmidrule(lr){2-7}
& AUC$\uparrow$
& AUC$\uparrow$
& AUC$\uparrow$
& AUC$\uparrow$
& RMSE$\downarrow$
& $\%$ \\
\midrule
Real
& $.927\pm.000$ & $.770\pm.005$ & $.926\pm.001$ & $.946\pm.001$ & $.423\pm.003$ & $0.0$ \\
\midrule
CTGAN
& $.886\pm.002$ & $.696\pm.005$ & $.875\pm.009$ & $.855\pm.006$ & $.902\pm.019$ & $28.48$ \\
TVAE
& $.878\pm.004$ & $.724\pm.005$ & $.871\pm.006$ & $.887\pm.003$ & $.770\pm.011$ & $21.09$ \\
GOGGLE
& $.778\pm.012$ & $.584\pm.005$ & $.658\pm.052$ & $.654\pm.024$ & $1.09\pm.025$ & $51.54$ \\
STaSy
& $.906\pm.001$ & $.752\pm.006$ & $.914\pm.005$ & $.934\pm.003$ & $.656\pm.014$ & $12.45$ \\
CoDi
& $.871\pm.006$ & $.525\pm.006$ & $.865\pm.006$ & $.932\pm.003$ & $.818\pm.021$ & $27.86$ \\
TabDDPM
& $.907\pm.001$ & $.758\pm.004$ & $.918\pm.005$ & $.935\pm.003$ & $.592\pm.011$ & $9.14$ \\
TabSyn
& $.909\pm.001$ & $.763\pm.002$ & $.914\pm.004$ & \redbf{$\mathbf{.937\pm.002}$} & $.580\pm.009$ & $8.44$ \\
TabDiff
& $.912\pm.002$ & $.763\pm.005$ & $.921\pm.004$ & $.936\pm.003$ & \redbf{$\mathbf{.555\pm.013}$} & \redbf{$\mathbf{7.07}$} \\
TabNAT
& $.904$ & $.764$ & $.916$ & $.935$ & $.579$ & $8.48$ \\
\midrule
GReaT
& $.913\pm.003$ & $.755\pm.006$ & $.902\pm.005$ & $.888\pm.008$ & $.653\pm.013$ & $13.31$ \\
DiffLM
& $.906$ & \redbf{$\mathbf{.794}$} & $.915$ & $.917$ & $.696$ & $13.59$ \\
Qwen2.5-7B$_{\text{ICL}}$ & $.853$ & $.390$ & $.811$ & $.829$ & $.991$ & $43.28$ \\
Qwen2.5-14B$_{\text{ICL}}$
& $.852$ & $.639$ & $.640$ & $.844$ & $1.03$ & $42.05$ \\
Qwen2.5-7B$_{\text{SFT}}$ & \redbf{$\mathbf{.915}$} & $.769$ & $.895$ & $.918$ & $.674$ & $13.41$ \\
LLaDA-8B$_{\text{SFT}}$ & $.909$ & $.774$ & $.872$ & $\mathbf{.919}$ & $.679$ & $14.13$ \\
\midrule
\textbf{\names} & $.908$ & $.786$ & $.908$ & $.915$ & $\mathbf{.651}$ & $\mathbf{11.82}$ \\
\textbf{\name} & $.907$ & $.791$ & \redbf{$\mathbf{.923}$} & $.905$ & $.696$ & $13.73$ \\
\bottomrule
\end{tabular}
\end{table*}

\subsection{Evaluation results of $\alpha$-Precision}
\label{apdx:alpha_precision}
We report $\alpha$-Precision results in Table~\ref{tab:alpha-precision}. \name achieves the best overall score on Default and the best average score among language-based models. Compared with tabular-specific synthetic models that are designed exclusively for numerical and categorical features, \name remains competitive while uniquely supporting free-form text generation.

\begin{table*}[htbp]
\centering
\footnotesize
\caption{Evaluation on $\alpha$-Precision scores. (\redbf{B}: best overall; \textbf{B}: best in language-based model)} 
\vspace{-5pt}
\setlength{\tabcolsep}{5pt}
\label{tab:alpha-precision}
\begin{tabular}{lccccc|>{\columncolor{gray!15}}c}
\toprule
\textbf{Method} & \textbf{Adult} & \textbf{Default} & \textbf{Shoppers} & \textbf{Magic} & \textbf{Beijing} & \textbf{Average} \\
\midrule
CTGAN   & $77.74\pm0.15$ & $62.08\pm0.08$ & $76.97\pm0.39$ & $86.90\pm0.22$ & $96.27\pm0.14$ & $79.99$ \\
TVAE    & $98.17\pm0.17$ & $85.57\pm0.34$ & $58.19\pm0.26$ & $86.19\pm0.48$ & $97.20\pm0.10$ & $85.06$ \\
GOGGLE  & $50.68$        & $68.89$        & $86.95$        & $90.88$        & $88.81$        & $77.24$ \\
STaSy   & $82.87\pm0.26$ & $90.48\pm0.11$ & $89.65\pm0.25$ & $86.56\pm0.19$ & $89.16\pm0.12$ & $87.74$ \\
CoDi    & $77.58\pm0.45$ & $82.38\pm0.15$ & $94.95\pm0.35$ & $85.01\pm0.36$ & $98.13\pm0.38$ & $87.61$ \\
TabDDPM & $96.36\pm0.20$ & $97.59\pm0.36$ & $88.55\pm0.68$ & $98.59\pm0.17$ & $97.93\pm0.30$ & $95.80$ \\
TabSyn  & \redbf{$\mathbf{99.39\pm0.18}$} & $98.65\pm0.23$ & $98.36\pm0.52$ & $99.42\pm0.28$ & $97.51\pm0.24$ & $98.67$ \\
TabDiff & $99.02\pm0.20$ & $98.49\pm0.28$ & \redbf{$\mathbf{99.11\pm0.34}$} & $99.47\pm0.21$ & $98.06\pm0.24$ & $98.83$ \\
TabNAT & $98.67$ & $99.27$ & $97.67$ & \redbf{$\mathbf{99.50}$} & \redbf{$\mathbf{99.27}$} & \redbf{$\mathbf{98.88}$} \\
\midrule
GReaT   & $55.79\pm0.03$ & $85.90\pm0.17$ & $78.88\pm0.13$ & $85.46\pm0.54$ & $\mathbf{98.32\pm0.22}$ & $80.87$ \\
Qwen2.5-7B$_{\text{ICL}}$ & $85.51$ & $88.56$ & $27.96$ & $83.39$ & $63.76$ & $69.84$ \\
Qwen2.5-14B$_{\text{ICL}}$ & $68.53$ & $83.17$ & $8.93$ & $89.99$ & $84.53$ & $67.03$ \\
Qwen2.5-7B$_{\text{SFT}}$ & $84.70$ & $95.25$ & $95.46$ & $82.45$ & $94.35$ & $90.44$ \\
LLaDA-8B$_{\text{SFT}}$ & $\mathbf{99.32}$ & $97.68$ & $68.73$ & $85.09$ & $95.13$ & $89.19$ \\
\midrule
\textbf{\names} & $96.51$ & $88.53$ & $97.23$ & $95.04$ & $97.62$ & $94.99$ \\
\textbf{\name} & $97.62$ & \redbf{$\mathbf{99.56}$} & $\mathbf{98.51}$ & $\mathbf{97.90}$ & $97.87$ & $\mathbf{98.29}$ \\
\bottomrule
\end{tabular}
\end{table*}

\begin{table*}[htbp]
\centering
\footnotesize
\caption{Evaluation on C2ST. (\redbf{B}: best overall; \textbf{B}: best in language-based model)} 
\vspace{-5pt}
\setlength{\tabcolsep}{10pt}
\label{tab:c2st}
\begin{tabular}{lccccc|>{\columncolor{gray!15}}c}
\toprule
\textbf{Method} & \textbf{Adult} & \textbf{Default} & \textbf{Shoppers} & \textbf{Magic} & \textbf{Beijing} & \textbf{Average} \\
\midrule
CTGAN   & $0.5949$ & $0.4875$ & $0.7488$ & $0.6728$ & $0.7531$ & $0.6514$ \\
TVAE    & $0.6315$ & $0.6547$ & $0.2962$ & $0.7706$ & $0.8659$ & $0.6438$ \\
GOGGLE  & $0.1114$ & $0.5163$ & $0.1418$ & $0.9526$ & $0.4779$ & $0.4400$ \\
STaSy   & $0.4054$ & $0.6814$ & $0.5482$ & $0.6939$ & $0.7922$ & $0.6242$ \\
CoDi    & $0.2077$ & $0.4595$ & $0.2784$ & $0.7206$ & $0.7177$ & $0.4768$ \\
TabDDPM & $0.9755$ & $0.9712$ & $0.8349$ & \redbf{$\mathbf{0.9998}$} & $0.9513$ & $0.9465$ \\
TabSyn  & $0.9910$ & \redbf{$\mathbf{0.9826}$} & $0.9662$ & $0.9960$ & $0.9528$ & $0.9777$ \\
TabDiff & \redbf{$\mathbf{0.9950}$} & $0.9774$ & \redbf{$\mathbf{0.9843}$} & $0.9989$ & $0.9781$ & \redbf{$\mathbf{0.9867}$} \\
TabNAT & $0.9870$ & $0.9657$ & $0.9626$ & $0.9989$ & \redbf{$\mathbf{0.9845}$} & $0.9797$ \\
\midrule
GReaT   & $0.5376$ & $0.4710$ & $0.4285$ & $0.4326$ & $0.6893$ & $0.5118$ \\
Qwen2.5-7B$_{\text{ICL}}$ & $0.0925$ & $0.3089$ & $0.0591$ & $0.5119$ & $0.1787$ & $0.2302$ \\
Qwen2.5-14B$_{\text{ICL}}$ & $0.1564$ & $0.4866$ & $0.0186$ & $0.5903$ & $0.2702$ & $0.3044$ \\
Qwen2.5-7B$_{\text{SFT}}$ & $0.8037$ & $0.9131$ & $0.5722$ & $0.7845$ & $0.7315$ & $0.7610$ \\
LLaDA-8B$_{\text{SFT}}$ & $0.9366$ & $0.9140$ & $0.5171$ & $0.8489$ & $\mathbf{0.9799}$ & $0.8393$ \\
\midrule
\textbf{\names} & $\mathbf{0.9491}$ & $0.7790$ & $0.8255$ & $\mathbf{0.9736}$ & $0.9527$ & $0.8960$ \\
\textbf{\name} & $0.9386$ & $\mathbf{0.9734}$ & $\mathbf{0.9594}$ & $0.9254$ & $0.9528$ & $\mathbf{0.9499}$ \\
\bottomrule
\end{tabular}
\vspace{-10pt}
\end{table*}

\subsection{Evaluation results of C2ST}
\label{apdx:c2st}
Table~\ref{tab:c2st} reports the C2ST detection score, where higher values indicate that synthetic samples are statistically harder to distinguish from real ones. \name achieves the best average result among all language-based models, improving the average C2ST score by $13.2\%$ compared with the strongest language-based baseline. Compared with tabular-specific diffusion models designed exclusively for numerical and categorical features, \name still achieves competitive performance.

The C2ST scores in Table~\ref{tab:c2st} follow the standard protocol with a logistic-regression discriminator. Table~\ref{tab:c2st_strong} repeats the test with two stronger, tuned discriminators, a random forest (RF) and gradient-boosted decision trees (GBDT), keeping the column set, encoding, folds, and scoring rule unchanged. Stronger discriminators lower the score of every system, and \name retains the highest average among language-model generators under all three.

\begin{table}[h]
\centering
\footnotesize
\setlength{\tabcolsep}{3.2pt}
\caption{Evaluation on C2ST using logistic regression (LR), random forest (RF), and gradient-boosted trees (GBDT) as the classifier.}
\label{tab:c2st_strong}
\begin{tabular}{lccccccccc|>{\columncolor{gray!15}}c>{\columncolor{gray!15}}c>{\columncolor{gray!15}}c}
\toprule
\multirow{2}{*}{\textbf{Method}} & \multicolumn{3}{c}{\textbf{Adult}} & \multicolumn{3}{c}{\textbf{Shoppers}} & \multicolumn{3}{c|}{\textbf{Default}} & \multicolumn{3}{>{\columncolor{gray!15}}c}{\textbf{Average}} \\
\cmidrule(lr){2-4}\cmidrule(lr){5-7}\cmidrule(lr){8-10}\cmidrule(lr){11-13}
& LR & RF & GBDT & LR & RF & GBDT & LR & RF & GBDT & LR & RF & GBDT \\
\midrule
Qwen2.5-7B$_{\text{ICL}}$ & $.093$ & $.004$ & $.001$ & $.059$ & $.010$ & $.005$ & $.309$ & $.006$ & $.001$ & $.153$ & $.007$ & $.002$ \\
Qwen2.5-14B$_{\text{ICL}}$ & $.156$ & $.013$ & $.004$ & $.019$ & $.000$ & $.000$ & $.487$ & $.004$ & $.001$ & $.221$ & $.006$ & $.002$ \\
Qwen2.5-7B$_{\text{SFT}}$ & $.804$ & $.612$ & $.560$ & $.572$ & $.314$ & $.248$ & $.913$ & \redbf{$\mathbf{.630}$} & \redbf{$\mathbf{.548}$} & $.763$ & $.519$ & $.452$ \\
LLaDA-8B$_{\text{SFT}}$ & $.937$ & $.737$ & \redbf{$\mathbf{.753}$} & $.517$ & $.132$ & $.089$ & $.914$ & $.564$ & $.547$ & $.789$ & $.478$ & $.463$ \\
\midrule
\textbf{\name} & \redbf{$\mathbf{.939}$} & \redbf{$\mathbf{.753}$} & $.744$ & \redbf{$\mathbf{.959}$} & \redbf{$\mathbf{.591}$} & \redbf{$\mathbf{.473}$} & \redbf{$\mathbf{.973}$} & $.532$ & $.418$ & \redbf{$\mathbf{.957}$} & \redbf{$\mathbf{.625}$} & \redbf{$\mathbf{.545}$} \\
\bottomrule
\end{tabular}
\end{table}

\subsection{Evaluation results of DCR}
\label{apdx:dcr}
Table~\ref{tab:dcr} reports the DCR score, which serves as a privacy metric: a value closer to $50\%$ indicates that synthetic samples are equally close to training and held-out test records, suggesting the model has learned the underlying distribution rather than memorized training instances. \name~achieves the best overall average deviation across all baselines and obtains the best per-dataset DCR on Default and Beijing. On average, the large pretrained backbone therefore does not lead to more memorization than specialized tabular diffusion models.

The DCR analysis above covers tables without free-form text. We extend it to Amazon. Following the same protocol, the real data is split into two equal disjoint halves, one used for training and one held out, and we report the fraction of synthetic rows that are closer to the training half; values closer to $50\%$ are better. Distances over numerical and categorical columns are computed as above; for the free-form text columns we use the cosine distance between term frequency--inverse document frequency (TF-IDF) vectors~\cite{salton1988term} computed over single words and pairs of consecutive words. As shown in Table~\ref{tab:dcr_text}, \name is the closest to $50\%$ on the structured channel and on average, whereas Qwen2.5-7B$_{\text{SFT}}$ is saturated on both channels, i.e., it largely reproduces its training set. Table~\ref{tab:ngram} additionally measures verbatim reuse, averaged over the three longest free-form text columns (\texttt{description}, \texttt{review\_text} and \texttt{summary}): the exact-duplicate rate, the fraction of generations whose TF-IDF similarity to a training text exceeds $0.8$, and the fraction of generated $n$-grams (runs of $n$ consecutive words) that occur verbatim in the training text. A held-out real split, scored identically, gives the level of reuse that is natural for this corpus. \name has the lowest long-span reuse among all fine-tuned systems.

\begin{table}[H]
\centering
\footnotesize
\caption{Evaluation on DCR score, where a score closer to $50\%$ is more preferable. We also report the average absolute deviation of the DCR score from $50\%$ across all datasets, where lower values are better. (\redbf{B}: best overall)}
\vspace{5pt}
\setlength{\tabcolsep}{6pt}
\label{tab:dcr}
\begin{tabular}{lcccc|>{\columncolor{gray!15}}c}
\toprule
\textbf{Method} & \textbf{Adult} & \textbf{Default} & \textbf{Shoppers} & \textbf{Beijing} & \textbf{Average $\downarrow$} \\
\midrule
STaSy   & $50.33\%\pm0.19$ & $50.23\%\pm0.09$ & $51.53\%\pm0.16$ & $50.59\%\pm0.29$ & $0.67\%$ \\
CoDi    & \redbf{$\mathbf{49.92\%\pm0.18}$} & $51.82\%\pm0.26$ & $51.06\%\pm0.18$ & $50.87\%\pm0.11$ & $0.96\%$ \\
TabDDPM & $51.14\%\pm0.18$ & $52.15\%\pm0.20$ & $63.23\%\pm0.25$ & $80.11\%\pm2.68$ & $11.66\%$ \\
TabSyn  & $50.94\%\pm0.17$ & $51.20\%\pm0.18$ & $52.90\%\pm0.22$ & $50.37\%\pm0.13$ & $1.35\%$ \\
TabDiff & $50.10\%\pm0.32$ & $51.11\%\pm0.36$ & \redbf{$\mathbf{50.24\%\pm0.62}$} & $50.50\%\pm0.36$ & $0.49\%$ \\
\midrule
\textbf{\name} & $50.73\%$& \redbf{$\mathbf{49.95\%}$} & $50.87\%$ & \redbf{$\mathbf{50.13\%}$} & \redbf{$\mathbf{0.45\%}$} \\
\bottomrule
\end{tabular}
\vspace{-10pt}
\end{table}

\begin{table}[H]
\centering
\footnotesize
\setlength{\tabcolsep}{6pt}
\caption{DCR on Amazon, separately for the structured columns and the free-form text columns. A score closer to $50\%$ is preferable.}
\vspace{5pt}
\label{tab:dcr_text}
\begin{tabular}{lccc}
\toprule
\textbf{Method} & \textbf{Num.+Cat.} & \textbf{Text} & \textbf{Mean $|\Delta$ from $50\%|$} $\downarrow$ \\
\midrule
Qwen2.5-7B$_{\text{ICL}}$ & $41.04\%$ & $37.93\%$ & $10.52\%$ \\
Qwen2.5-14B$_{\text{ICL}}$ & $44.68\%$ & \redbf{$\mathbf{42.01\%}$} & $6.66\%$ \\
Qwen2.5-7B$_{\text{SFT}}$ & $98.16\%$ & $99.85\%$ & $49.01\%$ \\
LLaDA-8B$_{\text{SFT}}$ & $48.05\%$ & $61.36\%$ & $6.66\%$ \\
\textbf{\name} & \redbf{$\mathbf{50.24\%}$} & $59.37\%$ & \redbf{$\mathbf{4.81\%}$} \\
\bottomrule
\end{tabular}
\end{table}

\begin{table}[H]
\centering
\footnotesize
\setlength{\tabcolsep}{5pt}
\caption{Verbatim reuse of training text on Amazon (lower is less reuse), averaged over the free-form text columns.}
\label{tab:ngram}
\begin{tabular}{lccccc}
\toprule
\textbf{Method} & \textbf{Exact dup.}\,$\downarrow$ & \textbf{Sim.\,$>$\,0.8}\,$\downarrow$ & \textbf{5-gram}\,$\downarrow$ & \textbf{8-gram}\,$\downarrow$ & \textbf{13-gram}\,$\downarrow$ \\
\midrule
Real held-out split (reference) & $29.78\%$ & $39.98\%$ & $13.96\%$ & $12.20\%$ & $11.51\%$ \\
\midrule
Qwen2.5-7B$_{\text{ICL}}$ & $23.28\%$ & \redbf{$\mathbf{8.00\%}$} & \redbf{$\mathbf{0.55\%}$} & $0.21\%$ & $0.15\%$ \\
Qwen2.5-14B$_{\text{ICL}}$ & $19.79\%$ & $8.99\%$ & $1.98\%$ & $0.22\%$ & $0.04\%$ \\
Qwen2.5-7B$_{\text{SFT}}$ & $36.26\%$ & $38.67\%$ & $9.76\%$ & $2.97\%$ & $2.07\%$ \\
LLaDA-8B$_{\text{SFT}}$ & $28.56\%$ & $33.96\%$ & $2.15\%$ & $0.18\%$ & $0.02\%$ \\
\textbf{\name} & \redbf{$\mathbf{18.27\%}$} & $24.04\%$ & $1.03\%$ & \redbf{$\mathbf{0.07\%}$} & \redbf{$\mathbf{0.00\%}$} \\
\bottomrule
\end{tabular}
\end{table}

\input{08b-additional}

\subsection{Description of Ablation Variants}
\label{apdx:ablation}
The variants in the ablation study are defined as follows. (i) \name-noFloatAE replaces the pretrained and frozen float encoder/decoder with randomly initialized modules of the same architecture, trained jointly with the rest of the model. (ii) \name-onlyContDiff converts categorical features into one-hot vectors and feeds them through the continuous diffusion branch, removing MDLM-based modeling for categorical fields. (iii) \name-onlyMDLM serializes numerical features and processes them in the same way as categorical and textual features, removing the continuous diffusion branch entirely. The aggregated results are reported in Table~\ref{tab:ablation}.

\input{05-limitation}

\section{Societal Impact}
\label{apdx:social_impact}
\name enables high-fidelity synthetic tabular data generation 
across numerical, categorical, and free-form text fields, which can 
support privacy-preserving data sharing, data augmentation in 
low-resource settings, and controlled benchmarks for tabular foundation 
model research. The DCR results in Appendix~\ref{apdx:dcr} suggest that 
\name learns the underlying distribution rather than 
memorizing training records, which is desirable for these use cases. 
As with any generative model, care should be taken when applying 
\name beyond the evaluated benchmarks. Synthetic rows that 
jointly couple structured attributes and fluent text could be misused 
for fabricated reviews or profiles, and generated data may inherit 
biases from the training distribution. High-stakes deployments should 
involve human oversight, privacy auditing, and bias evaluation on both 
the source data and the generated samples.

\stopcontents[appendix]

%% file: 08b-additional.tex
\subsection{Distributions of the numerical columns}
\label{apdx:marginals}
\begin{figure}[H]
\centering
\includegraphics[width=\linewidth]{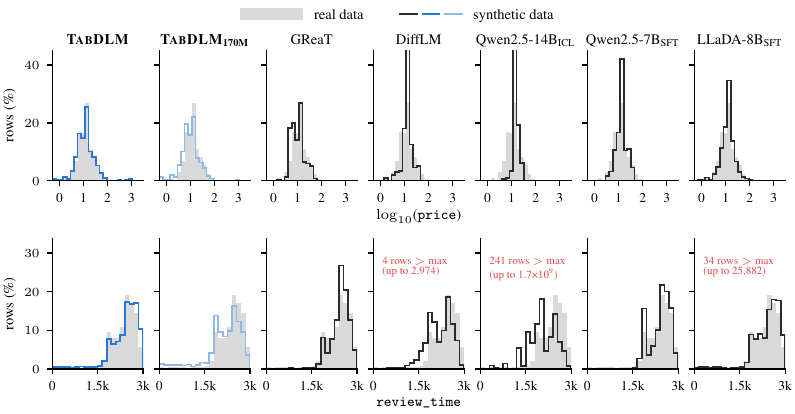}
\caption{Distributions of the numerical columns on Amazon (real maximum: \texttt{price} 1,000, \texttt{review\_time} 2,922). Red notes count synthetic values above the real maximum; bars taller than the axis are cut at the top.}
\label{fig:marg_amazon}
\end{figure}

\begin{figure}[H]
\centering
\includegraphics[width=\linewidth]{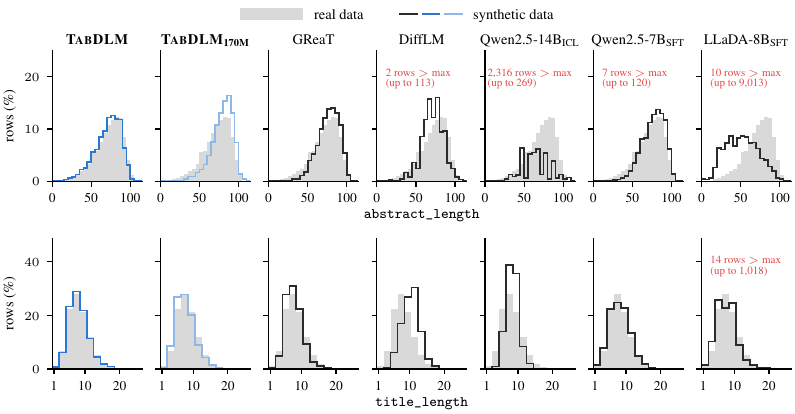}
\caption{Distributions of the numerical columns on Arxiv (real maximum: \texttt{abstract\_length} 109, \texttt{title\_length} 27), as in Figure~\ref{fig:marg_amazon}.}
\label{fig:marg_arxiv}
\end{figure}

Figures~\ref{fig:marg_amazon} and~\ref{fig:marg_arxiv} compare the distribution of every numerical column of Amazon and Arxiv in the generated rows with that in the real data. On the long-tailed \texttt{price} of Amazon, DiffLM and the Qwen2.5 baselines cut the tail: none of them generates a \texttt{price} above $66$, versus $1{,}000$ in the real data, whereas \name covers the full range of the real prices. On the other three columns most systems, including GReaT, are close to the real data, but some language-model baselines generate values far beyond the real maximum: $58\%$ of the \texttt{abstract\_length} values generated by Qwen2.5-14B$_{\text{ICL}}$ exceed the real maximum of $109$, and LLaDA-8B$_{\text{SFT}}$, which shares the backbone of \name but writes numbers as digit tokens, generates \texttt{abstract\_length} values of up to $9{,}013$. Neither variant of \name generates a value outside the real range on any of the four columns.

\subsection{Quality of the generated free-form text}
\label{apdx:textq}
\begin{figure}[H]
\centering
\includegraphics[width=0.78\linewidth]{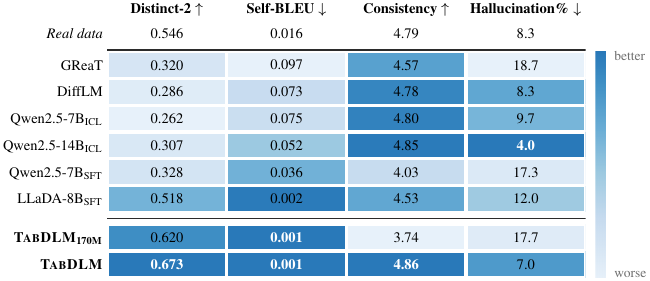}
\caption{Quality of the generated free-form text on Amazon. Consistency is the LLM-judge score (1--5); Hallucination\% is the percentage of rows whose text contradicts a numerical or categorical column. Within each column, bluer is better and the best value is in bold.}
\label{fig:textq}
\end{figure}

We evaluate the free-form text of the generated Amazon rows with four metrics. Distinct-2 (the fraction of unique bigrams) and Self-BLEU measure the diversity of the generated text. An LLM judge (\texttt{gpt-5-mini}) rates each row on a 1--5 scale for cross-field consistency and lists every statement in a text field that contradicts a numerical or categorical column of the same row; the hallucination rate is the fraction of rows with at least one such contradiction. Rows from all systems and real rows are pooled, shuffled and judged blind, five rows per request. As shown in Figure~\ref{fig:textq}, \name attains the highest Distinct-2 and judged consistency and the lowest Self-BLEU (tied with \names).

The judge prompt is given below: the system message, then the user message with the rubric and the rows to judge.

\clearpage
\begin{tcolorbox}[enhanced, colback=black!3, colframe=black!72, colupper=black, coltitle=white, fonttitle=\small, title={LLM-Judge Prompt}, arc=1.5mm, boxrule=0.6pt, left=2.5mm, right=2.5mm, top=1.5mm, bottom=1.5mm, fontupper=\footnotesize]
\setlength{\parindent}{0pt}\setlength{\parskip}{2pt}
\textbf{System}\\
You are a careful data-quality annotator for a table of Amazon product reviews.

Your job is to judge whether the fields of each row are mutually consistent and whether the free-text fields are grounded in the structured fields. You are NOT judging whether the review is well-written, helpful, or a product you would buy.

Critically: real Amazon reviews are often terse, ungrammatical, lowercase, off-topic, or oddly punctuated, and real ratings are sometimes a poor match for the review's tone. None of that is a defect here. Penalize only genuine cross-field contradictions and genuine incoherence -- never mere brevity, informality, or an unremarkable product.

Judge only what is present in the row. Do not use outside knowledge about whether a product or brand really exists.

Return only JSON matching the provided schema.

\textbf{User}\\
Each row is one product review. Fields:\\[1pt]
\begin{tabular}{@{\hspace{1.2em}}l@{\hspace{1em}}l@{}}
\texttt{price} & numeric product price\\
\texttt{rating} & integer 1--5, the reviewer's star rating\\
\texttt{category} & product category\\
\texttt{verified} & whether the purchase was verified\\
\texttt{brand}, \texttt{title} & the product's brand and product title\\
\texttt{description} & the product's description text\\
\texttt{review\_text} & the reviewer's free-text review\\
\texttt{summary} & the reviewer's short headline for the review
\end{tabular}

Score each dimension 1--5 (integers only).

1) \texttt{rating\_consistency} -- does the sentiment of \texttt{review\_text}/\texttt{summary} match \texttt{rating}?\\[1pt]
\begin{tabular}{@{\hspace{1.2em}}l@{\hspace{0.8em}}p{0.86\linewidth}@{}}
5 & clearly matches (5 stars with praise; 1 star with complaints; 3 stars mixed)\\
4 & matches, with minor ambiguity\\
3 & unclear or genuinely mixed; no outright contradiction\\
2 & probable mismatch (rating 5, review mostly negative)\\
1 & flat contradiction (rating 5 with ``broke immediately, do not buy'')
\end{tabular}\\
\hspace*{1.2em}If \texttt{review\_text} is empty or contentless, score 3 and say so -- do not guess.

2) \texttt{product\_consistency} -- do brand / title / description / category describe ONE coherent product, and does \texttt{review\_text} review THAT product?\\[1pt]
\begin{tabular}{@{\hspace{1.2em}}l@{\hspace{0.8em}}p{0.86\linewidth}@{}}
5 & all fields point to the same product\\
4 & coherent; one field vague or generic\\
3 & some drift, still plausibly the same product\\
2 & a field clearly describes a different product type\\
1 & fields are unrelated (title a kitchen tool, review about a paperback)
\end{tabular}

3) \texttt{summary\_entailment} -- is \texttt{summary} supported by \texttt{review\_text}?\\[1pt]
\begin{tabular}{@{\hspace{1.2em}}l@{\hspace{0.8em}}p{0.86\linewidth}@{}}
5 & summary is a faithful compression of the review\\
4 & faithful, slightly broader or narrower\\
3 & related but adds or omits something notable\\
2 & summary asserts something the review does not support\\
1 & summary contradicts the review, or describes a different experience
\end{tabular}\\
\hspace*{1.2em}If either field is empty, score 3.

4) \texttt{intra\_text\_coherence} -- are description / review\_text / summary each internally coherent as writing?\\[1pt]
\begin{tabular}{@{\hspace{1.2em}}l@{\hspace{0.8em}}p{0.86\linewidth}@{}}
5 & reads as a real human-written passage\\
4 & minor awkwardness\\
3 & noticeably odd but followable\\
2 & disjointed, self-contradicting, or looping\\
1 & degenerate: repeated n-grams, truncated mid-word, word salad
\end{tabular}\\
\hspace*{1.2em}Terse-but-sensible text is a 5, not a 2. Length is not a defect.

5) \texttt{hallucinations} -- list every SPECIFIC assertion in the text fields that CONTRADICTS a structured field of this same row. For each, quote the span verbatim and name the column it contradicts.\\
\hspace*{1.2em}\parbox[t]{\dimexpr\linewidth-1.2em\relax}{Examples that count: the review says ``for the price of under \$5'' while price is 89.99; the review says ``this Nikon lens'' while brand is Logitech; the review says ``arrived in the Kindle store'' while category is Pet Supplies. Does NOT count: any detail that is merely unverifiable (an invented model number, a made-up anecdote, a fictional-sounding brand). Unverifiable is not contradiction -- this list must stay checkable, so contradiction with a column in this row is the only criterion. Empty list is the common case.}

Rows to judge (each a JSON object). They come from different sources in no particular order; judge each on its own.

\{one JSON object per row: \texttt{row\_id}, \texttt{price}, \texttt{rating}, \texttt{category}, \texttt{verified}, \texttt{brand}, \texttt{title}, \texttt{description}, \texttt{review\_text}, \texttt{summary}\}

Return JSON with key \texttt{results}: exactly 5 elements, one per \texttt{row\_id} above.

\textbf{Assistant}
\end{tcolorbox}
\clearpage

\subsection{Case study}
\label{apdx:examples}
The boxes below show randomly sampled rows generated by each system on Arxiv. In the real data, \texttt{title\_length} and \texttt{abstract\_length} are exactly the numbers of words in \texttt{title} and \texttt{abstract}, and \texttt{category} is the primary arXiv category of the paper, so all three can be checked against the text of the same row. Values that contradict the text of their row or lie outside the range of the real data, as well as repeated or malformed passages of the text, are marked in red and explained below the row; orange marks a minor deviation.

\begin{tcolorbox}[exours, title={\name}]
\exfields{$891$}{$2020$}{$5$}{$77$}{\texttt{gr-qc}}{\texttt{arXiv:2006.09404}}
\extitle{Quantum Charges around Static Spheres}
\exabstract{We introduce a new effective formalism to calculate the potential seen by quantum fields around a static object. First proposed by Schwinger, this approach refers to a Minkowski reference space and provides a ready-to-use calculation of the potential using any quantum field theory. We show that within the framework of asymptotic safety, one can deduce the gravitational potential up to second order in the metric perturbation from this potential. A numerical illustration of our method is provided.}
\exnotes{\exok \texttt{title\_length} and \texttt{abstract\_length} match the title ($5$ words) and the abstract ($77$ words), and the paper fits its \texttt{category}.}
\exsep
\exfields{$1405$}{$2021$}{$6$}{$71$}{\texttt{quant-ph}}{\texttt{arXiv:2101.00211}}
\extitle{Causal effects of interacting quantum systems}
\exabstract{We consider mutually entangled quantum subsystems whose total state is repeatedly prepared and measured, and they are taken to be dynamically relaxing down to an asymptotic steady state where measurements yield all coherent contributions. We show that the dynamics of at most one subsystem makes the dynamics dependent on the trajectory of another subsystem, and this dependence persists when conditioned on measurement. These results demonstrate ``causal effects'' of interacting quantum systems.}
\exnotes{\exok \texttt{title\_length} and \texttt{abstract\_length} match the title ($6$ words) and the abstract ($71$ words), and the paper fits its \texttt{category}.}
\end{tcolorbox}

\begin{tcolorbox}[exourslight, title={\names}]
\exfields{$973$}{$2020$}{$4$}{\exminor{$89$}}{\texttt{hep-th}}{\texttt{arXiv:2009.06098}}
\extitle{Effective D variable analysis}
\exabstract{We study the static D field equations in the cosmological Nambu--Walker spacetime with arbitrary proper time, having consequences which are not unitary local. When the boundary of D fields is changed, a directly interacting Newtonian term emerges appearing in the resulting action that can be removed, by a remarkable operator dependent hypergeometric transformation. As fixed points is expanded, proper-time CPT disappears, which leads to vanishing of the additional term. The resulting action preserves all conformal dependence and hence provide a direct, unitary, solution with no trace of nonlocal boundary conditions.}
\exnotes{\exok \texttt{title\_length} matches the title ($4$ words), and the paper fits its \texttt{category}.\\
\exmin \texttt{abstract\_length} is $89$, one word fewer than the abstract ($90$ words).}
\end{tcolorbox}

\begin{tcolorbox}[exbase, title={GReaT}]
\exfields{$1625$}{$2022$}{\exbad{$8$}}{$95$}{\texttt{hep-ph}}{\texttt{arXiv:2207.08231}}
\extitle{Black Hole Thermodynamics in D=2}
\exabstract{We consider the thermodynamic properties of a two-dimensional black hole in the framework of the D=2 conformal field theory. In the first solution, the thermodynamic entropy is obtained by the Ramsey scheme, while in the second solution it depends on the thermodynamic entropy. Under the Ramsey scheme, the entropy becomes a dependent quantity in the thermodynamic field theory. \exbadtxt{Under the inverted symmetries, the thermodynamic entropy, under the inverted symmetries, depends on the parameters of the inverted symmetries. Under the inverted symmetries, the thermodynamic entropy depends on the parameters of the inverted symmetries.}}
\exnotes{\exno \texttt{title\_length} is $8$, but the title has $5$ words.\\
\exno The last two sentences say the same thing and repeat ``the inverted symmetries'' five times.}
\end{tcolorbox}

\begin{tcolorbox}[exbase, title={DiffLM}]
\exfields{$1655$}{$2022$}{\exbad{$13$}}{\exbad{$60$}}{\texttt{quant-ph}}{\texttt{arXiv:2206.11744}}
\extitle{A new type of quantum error correction for quantum networks}
\exabstract{Quantum error correction is a fundamental tool for error-free quantum communication in a quantum network. We propose a new type of quantum error correction scheme, which is based on the quantum teleportation protocol, and can be used to correct errors in quantum networks. The scheme is simple, practical, and can be easily extended to correct errors in quantum networks with more than two nodes. \exbadtxt{The proposed scheme can be used to correct errors in a quantum network with only two nodes, and it can also be used to correct errors in a quantum network with more than two nodes.}}
\exnotes{\exno \texttt{title\_length} is $13$ and \texttt{abstract\_length} is $60$, but the title has $10$ words and the abstract $99$.\\
\exno The last sentence repeats the previous one.}
\end{tcolorbox}

\begin{tcolorbox}[exbase, title={Qwen2.5-7B$_{\text{ICL}}$}]
\exfields{$365$}{\exbad{$2023$}}{\exbad{$12$}}{\exbad{$150$}}{\exbad{\texttt{astro-ph.SR}}}{\exbad{\texttt{arXiv:2303.12345}}}
\extitle{\exbadtxt{Exploring Quantum Entanglement in Black Hole Horizons}}
\exabstract{This paper delves into the quantum entanglement properties near black hole horizons, presenting new theoretical models that predict enhanced entanglement due to Hawking radiation. The study uses advanced computational techniques to simulate these effects and discusses potential implications for quantum gravity theories.}
\exnotes{\exno \texttt{title\_length} is $12$ and \texttt{abstract\_length} is $150$, but the title has $7$ words and the abstract $42$; $150$ also exceeds the longest real abstract ($109$ words).\\
\exno \texttt{submission\_year} is $2023$, outside the years of the real data ($2018$--$2022$).\\
\exno \texttt{category} is \texttt{astro-ph.SR} (solar and stellar astrophysics), but the paper is about quantum entanglement at black-hole horizons.\\
\exno The same title appears in $241$ and the same \texttt{arxiv\_code} in $597$ of the $4{,}000$ generated rows.}
\end{tcolorbox}

\begin{tcolorbox}[exbase, title={Qwen2.5-14B$_{\text{ICL}}$}]
\exfields{$365$}{$2022$}{\exbad{$9$}}{\exbad{$120$}}{\texttt{cond-mat.mes-hall}}{\exbad{\texttt{arXiv:2208.01234}}}
\extitle{\exbadtxt{Exploring Quantum Entanglement in Graphene Bilayers}}
\exabstract{Recent theoretical studies predict that quantum entanglement can significantly enhance the electronic properties of graphene bilayers. This paper explores these predictions through advanced computational models, providing insights into the potential applications in quantum computing and nanotechnology.}
\exnotes{\exno \texttt{title\_length} is $9$ and \texttt{abstract\_length} is $120$, but the title has $6$ words and the abstract $36$; $120$ also exceeds the longest real abstract ($109$ words).\\
\exno The same title appears in $200$ of the $4{,}000$ generated rows, and $767$ generated \texttt{arxiv\_code} values end in \texttt{01234}.}
\end{tcolorbox}

\begin{tcolorbox}[exbase, title={Qwen2.5-7B$_{\text{SFT}}$}]
\exfields{$1117$}{$2021$}{\exbad{$6$}}{\exbad{$90$}}{\exbad{\texttt{hep-th}}}{\texttt{arXiv:2004.07125}}
\extitle{Perturbative effects of a magnetic field on the dispersion relations of Bloch electrons}
\exabstract{We study the effect of a magnetic field on the dispersion relations of electrons in a two-dimensional lattice under the influence of a generalized Peierls--Aharonov and Berry phases. The latter is constructed from the Berry connection of Bloch electrons in the high symmetry points of the Brillouin zone. We analyze the perturbative effects of magnetic field on the dispersion relations of electron states and find the conditions for the source of the topological states at the zone boundary.}
\exnotes{\exno \texttt{title\_length} is $6$ and \texttt{abstract\_length} is $90$, but the title has $13$ words and the abstract $78$.\\
\exno \texttt{category} is \texttt{hep-th} (high-energy physics theory), but the paper is about electrons in a crystal lattice, a condensed-matter topic.}
\end{tcolorbox}

\begin{tcolorbox}[exbase, title={LLaDA-8B$_{\text{SFT}}$}]
\exfields{$35$}{$2018$}{\exbad{$9$}}{\exbad{$50$}}{\exbad{\texttt{quant-ph}}}{\texttt{arXiv:1802.01802}}
\extitle{Coordinate Independence and Dimension Dimensionality in \exbadtxt{FriedmannEquations}}
\exabstract{In the present work, we explore the relation between the generalized dimensions and generalized anisotropic space model in the Friedmann Equation. The solution of the generalized Friedmann equation and the generalized and the anisotropic Friedmann equations are obtained. We also find the anisotropic and generalized static solutions for the static case for both the generalized and anisotropic Friedmann equations.}
\exnotes{\exno \texttt{title\_length} is $9$ and \texttt{abstract\_length} is $50$, but the title has $7$ words and the abstract $59$.\\
\exno Two words of the title are run together.\\
\exno \texttt{category} is \texttt{quant-ph} (quantum physics), but the paper is about the Friedmann equations of cosmology.}
\end{tcolorbox}

In both rows of \name, \texttt{title\_length} and \texttt{abstract\_length} match the text exactly, the paper fits its \texttt{category}, and the abstract is coherent and free of repetition. \names makes one minor slip, an \texttt{abstract\_length} one word shorter than its abstract, which we attribute to its much smaller backbone. Every baseline row, in contrast, contains at least one clear error. The most common is a length value that contradicts the text of its own row: Qwen2.5-7B$_{\text{ICL}}$ reports a $150$-word abstract for an abstract of $42$ words, and LLaDA-8B$_{\text{SFT}}$, which shares the backbone of \name but writes numbers as digit tokens, misreports both lengths, in line with Figure~\ref{fig:arxiv_agree}. The text of the baselines degrades as well: GReaT and DiffLM repeat the same sentence, and the in-context baselines fall back on a few templates, reusing the same title in hundreds of rows.

%% file: 05-limitation.tex
\section{Limitations and Future Directions}\label{sec:limitations}
First, the sampling efficiency of \name is lower than that of existing tabular data generation methods such as TabDiff~\cite{tabdiff} and TabSyn~\cite{zhang2024mixedtype}. This limitation can be primarily attributed to the large model size of the MDLM backbone. Nevertheless, \name is designed for broader application scenarios, as it supports the generation of numerical, categorical, and free-form text fields, which specialized tabular diffusion models do not support. Moreover, common methods for accelerating MDLM inference, like KV caching~\cite{wu2025fast}, could be incorporated to improve sampling efficiency. However, it is out of the scope of this work. Second, \name models numerical and language diffusion using a shared noise schedule that couples the denoising dynamics across modalities and enforces a common step size for denoising both numerical and textual features. While effective in practice, this design may be suboptimal in certain scenarios. Introducing modality-specific noise schedules to partially decouple the denoising processes can be promising, which we leave for future investigation.

%% file: 09-checklist.tex
\section*{NeurIPS Paper Checklist}

\begin{enumerate}

\item {\bf Claims}
    \item[] Question: Do the main claims made in the abstract and introduction accurately reflect the paper's contributions and scope?
    \item[] Answer: \answerYes{}
    \item[] Justification: The claims in the abstract and introduction accurately reflect our contributions and scope.

\item {\bf Limitations}
    \item[] Question: Does the paper discuss the limitations of the work performed by the authors?
    \item[] Answer: \answerYes{}
    \item[] Justification: Limitations are discussed in Appendix~\ref{sec:limitations}.

\item {\bf Theory assumptions and proofs}
    \item[] Question: For each theoretical result, does the paper provide the full set of assumptions and a complete (and correct) proof?
    \item[] Answer: \answerNA{}
    \item[] Justification: This paper does not include theoretical results.

    \item {\bf Experimental result reproducibility}
    \item[] Question: Does the paper fully disclose all the information needed to reproduce the main experimental results of the paper to the extent that it affects the main claims and/or conclusions of the paper (regardless of whether the code and data are provided or not)?
    \item[] Answer: \answerYes{}
    \item[] Justification: The model architecture, training, and sampling procedures are described in Section~\ref{sec:methods}, with full implementation details and hyperparameters in Appendix~\ref{apdx:impl} and dataset construction in Appendix~\ref{apdx:dataset}.

\item {\bf Open access to data and code}
    \item[] Question: Does the paper provide open access to the data and code, with sufficient instructions to faithfully reproduce the main experimental results, as described in supplemental material?
    \item[] Answer: \answerYes{}
    \item[] Justification: Code is released at the URL provided in Appendix~\ref{apdx:impl}. Synthetic dataset construction is fully specified in Appendix~\ref{apdx:dataset}; all real-world datasets are publicly available.

\item {\bf Experimental setting/details}
    \item[] Question: Does the paper specify all the training and test details (e.g., data splits, hyperparameters, how they were chosen, type of optimizer) necessary to understand the results?
    \item[] Answer: \answerYes{}
    \item[] Justification: Training and evaluation details are provided in  Appendix~\ref{apdx: experiment} and Appendix~\ref{apdx:impl}.

\item {\bf Experiment statistical significance}
    \item[] Question: Does the paper report error bars suitably and correctly defined or other appropriate information about the statistical significance of the experiments?
    \item[] Answer: \answerYes{}
    \item[] Justification: We report mean $\pm$ standard deviation over multiple runs for baselines reproduced from prior work (Appendix~\ref{apdx:full_shape_trend}). For language-model-based methods, we use a single fixed random seed shared across all methods due to compute cost, as documented in Appendix~\ref{apdx:impl}.

\item {\bf Experiments compute resources}
    \item[] Question: For each experiment, does the paper provide sufficient information on the computer resources (type of compute workers, memory, time of execution) needed to reproduce the experiments?
    \item[] Answer: \answerYes{}
    \item[] Justification: All experiments are run on a single NVIDIA A100 GPU (80GB), with per-dataset training epochs reported in Appendix~\ref{apdx:impl}.
    
\item {\bf Code of ethics}
    \item[] Question: Does the research conducted in the paper conform, in every respect, with the NeurIPS Code of Ethics \url{https://neurips.cc/public/EthicsGuidelines}?
    \item[] Answer: \answerYes{}
    \item[] Justification: The research conforms to the NeurIPS Code of Ethics.

\item {\bf Broader impacts}
    \item[] Question: Does the paper discuss both potential positive societal impacts and negative societal impacts of the work performed?
    \item[] Answer: \answerYes{}
    \item[] Justification: Both positive and negative societal impacts, along with mitigations, are discussed in Appendix~\ref{apdx:social_impact}.
    
\item {\bf Safeguards}
    \item[] Question: Does the paper describe safeguards that have been put in place for responsible release of data or models that have a high risk for misuse (e.g., pre-trained language models, image generators, or scraped datasets)?
    \item[] Answer: \answerYes{}
    \item[] Justification: We evaluate memorization risk via the DCR metric 
(Appendix~\ref{apdx:dcr}), and ProfileBio is constructed without sensitive personal attributes (Appendix~\ref{apdx:profbio}). Further safeguards are discussed in Appendix~\ref{apdx:social_impact}.

\item {\bf Licenses for existing assets}
    \item[] Question: Are the creators or original owners of assets (e.g., code, data, models), used in the paper, properly credited and are the license and terms of use explicitly mentioned and properly respected?
    \item[] Answer: \answerYes{}
    \item[] Justification: All datasets and pretrained models are cited to their original sources and used in accordance with their public licenses.

\item {\bf New assets}
    \item[] Question: Are new assets introduced in the paper well documented and is the documentation provided alongside the assets?
    \item[] Answer: \answerYes{}
    \item[] Justification: The two new synthetic datasets, MathExpr and ProfileBio, are documented in Appendix~\ref{apdx:mathexpr} and~\ref{apdx:profbio}. Source code is released at the URL in Appendix~\ref{apdx:impl}.

\item {\bf Crowdsourcing and research with human subjects}
    \item[] Question: For crowdsourcing experiments and research with human subjects, does the paper include the full text of instructions given to participants and screenshots, if applicable, as well as details about compensation (if any)? 
    \item[] Answer: \answerNA{}
    \item[] Justification: This work does not involve crowdsourcing or human subjects research.

\item {\bf Institutional review board (IRB) approvals or equivalent for research with human subjects}
    \item[] Question: Does the paper describe potential risks incurred by study participants, whether such risks were disclosed to the subjects, and whether Institutional Review Board (IRB) approvals (or an equivalent approval/review based on the requirements of your country or institution) were obtained?
    \item[] Answer: \answerNA{}
    \item[] Justification: This work does not involve human subjects research.

\item {\bf Declaration of LLM usage}
    \item[] Question: Does the paper describe the usage of LLMs if it is an important, original, or non-standard component of the core methods in this research? Note that if the LLM is used only for writing, editing, or formatting purposes and does \emph{not} impact the core methodology, scientific rigor, or originality of the research, declaration is not required.
    \item[] Answer: \answerYes{}
    \item[] Justification: \name is built on a pretrained masked diffusion language model (LLaDA-8B), as described in Section~\ref{sec:methods} and Appendix~\ref{apdx:impl}. The text-quality evaluation in Appendix~\ref{apdx:textq} uses an LLM judge (\texttt{gpt-5-mini}), whose full prompt is given there.

\end{enumerate}